%% file: main.tex
\documentclass[journal]{IEEEtran}
\usepackage{amsmath,amsfonts}
\usepackage{algorithmic}
\usepackage{array}
\usepackage[caption=false,font=normalsize,labelfont=sf,textfont=sf]{subfig}
\usepackage{textcomp}
\usepackage{stfloats}
\usepackage{url}
\usepackage{verbatim}
\usepackage{graphicx}
\usepackage{titlesec}
\usepackage{indentfirst}

\titlespacing*{\section}{0pt}{2ex plus .5ex minus .1ex}{1ex plus .1ex}
\titlespacing*{\subsection}{0pt}{1ex plus .5ex minus .1ex}{0.3ex plus .1ex}

\titleformat{\subsubsection}[runin]
  {\normalfont\normalsize\itshape}{\arabic{subsubsection})}{0.5em}{}
\titlespacing*{\subsubsection}{0pt}{1ex plus .3ex minus .1ex}{0.5em}

\usepackage[T1]{fontenc}
\usepackage{lettrine}

\usepackage{ragged2e}

\usepackage[utf8]{inputenc}
\usepackage{microtype}
\usepackage{cite}
\usepackage{amsmath}
\usepackage{booktabs}
\usepackage{multirow}
\usepackage{tabularx}
\usepackage{makecell}
\newtheorem{definition}{Definition}

\usepackage[hidelinks]{hyperref}
\newcolumntype{Y}{>{\raggedright\arraybackslash}X}

\usepackage{listings}
\usepackage{tcolorbox}
\usepackage{soul,color}
\tcbuselibrary{skins,listings,breakable}
\lstdefinestyle{sigstyle}{
  basicstyle=\ttfamily\footnotesize,
  columns=fullflexible,
  keepspaces=true,
}

\begin{document}

\title{Knowledge Graph-Enhanced Zero-Shot Topic Classification: A Multi-Strategy Comparative Study \\
\vspace{0.5em}

}

\author{Shahana Akter,~Yatharth~Vohra,~Ankita~Shukla,~and~Souvika~Sarkar%
\thanks{S. Akter and S. Sarkar are with the A2I Lab, School of Computing, Wichita State University, Wichita, KS, USA (e-mail: sxakter7@shockers.wichita.edu; souvika.sarkar@wichita.edu).}%
\thanks{Y. Vohra and A. Shukla are with the MAGI Lab, College of Engineering, University of Nevada, Reno, Reno, NV, USA (e-mail: yvohra@unr.edu; ankitas@unr.edu).}}


\maketitle

\input{latex/Sections/abstract}
\input{latex/Sections/introduction}
\input{latex/Sections/related_work}
\input{latex/Sections/problem_statement}
\input{latex/Sections/methodology}
\input{latex/Sections/experimental_setup}

\input{latex/Sections/results_and_analysis}

\input{latex/Sections/conclusion}
\input{latex/Sections/limitation}
\input{latex/Sections/acknowledgement}

\bibliographystyle{IEEEtran}
\bibliography{latex/reference}
\input{latex/Sections/bib}

\end{document}

%% file: latex/Sections/abstract.tex
\begin{abstract}

Multi-label topic classification without labeled training data remains a challenging problem, particularly for documents containing complex semantic and relational information. In this paper, we introduce a knowledge graph-enhanced inference framework for zero-shot multi-label topic classification. The proposed approach enriches document representations with document-specific knowledge graphs constructed from subject-predicate-object triples, enabling models to explicitly leverage relational information during topic inference without requiring task-specific training data. To evaluate the effectiveness of this approach, we compare graph-enhanced and non-enhanced inference variants under keyword-guided prompting and self-consistency decoding strategies across 15 LLMs and 8 multi-label datasets spanning diverse domains. Experimental results show that knowledge graph-enhanced inference improves the performance of smaller LLMs, enabling more models to outperform a strong baseline, while providing limited or negative benefits for larger models, suggesting that these models already capture substantial relational knowledge during pretraining. In contrast, self-consistency decoding does not yield performance gains despite increasing computational costs by approximately fivefold. Our findings provide new insights into the role of explicit relational knowledge in zero-shot multi-label classification and offer practical guidance on when knowledge graph-enhanced inference is beneficial for LLM-based topic prediction.

\end{abstract}

%% file: latex/Sections/introduction.tex
\section{Introduction}
\lettrine{T}{ext} classification, the task of assigning one or more labels to a document based on its content, is a foundational problem in natural language processing~\cite{alghamdi2015survey,chauhan2021topic}. Large language models (LLMs) have reshaped this task by enabling zero-shot and few shot classification directly from natural language instructions, without task specific training data~\cite{brown2020language,yin2019benchmarking}, a capability particularly valuable when labeled data is scarce or label sets must be defined dynamically at inference time. A challenging variant is \emph{multi-label} classification, where a document may belong to several categories simultaneously~\cite{veeranna2016semantic}: a health article may address both \textit{Heart Health} and \textit{Women's Health}, while a product may discuss both \textit{Lens} and \textit{Battery} performance. This challenge is compounded in zero-shot settings, where the model must assign previously unseen or user defined labels relying solely on its pretrained knowledge and the semantics of the label names themselves.

Recent work has shown that zero-shot multi-label topic inference is feasible using either sentence encoder similarity or prompt based LLM classification~\cite{sarkar2023zeroshot}. In such settings, users provide a collection of documents together with a set of candidate topic labels, and optionally short keyword descriptions for each topic, and the system must assign zero or more labels to each document~\cite{sarkar2023zeroshot,reimers2019sentence}. This progress, however, has largely overlooked how documents and labels are internally represented during inference. Most existing zero-shot approaches represent documents and labels as largely flat text objects \cite{sarkar2022exploring, puri2019zero, yin2019benchmarking}. Sentence encoder methods \cite{reimers2019sentence, sarkar2022exploring} collapse a document into a single embedding vector, while direct prompting \cite{puri2019zero, kojima2022large} methods typically supply only the raw article text and a list of candidate labels or keywords. In both cases, the relational structure inside the document remains implicit. As a result, documents that share overlapping vocabulary but differ in how underlying concepts relate to one another can be difficult to distinguish~\cite{sarkar2023zeroshot}. For example, the topics \textit{Mental Health} and \textit{Brain and Cognitive Health} may be associated with highly similar terms, even when the underlying conceptual relationships expressed in the document differ in meaningful ways.

Knowledge graphs \cite{hogan2021knowledge} offer a natural and principled solution to this problem, as they represent information not as flat text but as subject predicate object triples \cite{ji2021survey} that make semantic relationships explicit rather than implicit. By exposing these relationships directly, a knowledge graph preserves precisely the structural information that flat text representations discard, information that becomes critical when documents share surface level vocabulary but diverge in underlying meaning. Injecting such relational information into the LLM inference process provides the model with explicit contextual grounding for its topic predictions. Motivated by this insight, our study systematically examines the impact of document level knowledge graphs on zero-shot multi-label topic classification. For each input document, we construct a knowledge graph using an LLM driven entity relationship extraction pipeline inspired by KGGen~\cite{mo2025kggen}. The resulting graph is then supplied to the LLM jointly with the input document at test time, allowing the model to reason over both the raw text and its extracted relational structure. Through this design, we systematically evaluate when explicit relational knowledge improves classification performance and how its effectiveness varies across LLMs of differing scale, offering, to our knowledge, the first systematic study of how document level knowledge graphs affect zero-shot multi-label classification across models of varying capacity.

To achieve a comprehensive and rigorous evaluation, we conduct an extensive exploratory study spanning fifteen LLMs (LLaMA 3.2-3B \cite{grattafiori2024llama}, Gemma 2-9B \cite{team2024gemma}, Qwen 2.5-72B\cite{yang2025qwen3} etc.), categorized into small (4B or less), medium (from 7B to 20B), and large (27B or more) groups based on parameter size, and eight multi-label datasets ~\cite{sarkar2023zeroshot, mohammad2018semeval} across diverse domains of applicability. We evaluate non graph and graph-enhanced inference under both keyword guided prompting and self-consistency decoding \cite{wang2022selfconsistency}, and benchmark all approaches against a strong baseline~\cite{sarkar2023zeroshot}. This broad experimental design allows us to draw several important, and at times counterintuitive, conclusions. Knowledge graph augmentation substantially improves the performance of smaller LLMs, enabling them to outperform the baseline~\cite{sarkar2023zeroshot} far more consistently than their non augmented counterparts. Larger LLMs, by contrast, benefit little from graph augmentation, suggesting that these models already encode much of the necessary relational knowledge during pre-training \cite{petroni2019language} and therefore have less need for it to be supplied explicitly at inference time. Graph augmentation is useful for small and mid-sized models in terms of precision with some sacrifice of recall. This implies that graph augmentation makes these models to be selective in their predictions. With large models, there are instances where graph augmentation leads to irrelevant connections, thereby making over-predictions. We further find that self-consistency decoding fails to improve classification performance, despite increasing inference cost fivefold, indicating that this widely used decoding strategy \cite{wang2022selfconsistency} offers limited practical value in this setting. Taken together, these findings offer new empirical insight into when and for whom explicit structured knowledge is beneficial, insight that, to our knowledge, has not been systematically established in prior zero-shot classification literature. Specifically, our contributions and key findings are as follows.

\begin{enumerate}
    \item We develop a novel knowledge graph-enhanced zero-shot inference framework for multi-label topic classification, which enriches document representations with knowledge graphs encoding subject-predicate-object relationships. (Section \ref{sec:methodoloy})

    \item We conduct a thorough exploratory study comparing graph-enhanced and graph-free inference under both keyword guided prompting and self-consistency decoding \cite{wang2022selfconsistency}, spanning fifteen LLMs and eight multi-label classification datasets, one of the most extensive evaluations of its kind to date. (Section \ref{sec:methodoloy})

    \item We show that explicit relational knowledge benefits smaller LLMs but provides no improvement, for larger LLMs, contributing new understanding of how model scale governs the utility of externally supplied structured knowledge. (Section \ref{sec:results})

    \item We find that medium sized models can occasionally outperform larger models. This suggests that increasing model size alone does not always necessarily lead to better zero-shot classification performance. (Section \ref{sec:results})

    \item We demonstrate that self-consistency decoding \cite{wang2022selfconsistency} substantially increases inference cost without yielding corresponding gains in classification performance which raises practical questions about its cost effectiveness in zero-shot classification pipelines. (Section \ref{sec:runtime})
\end{enumerate}

%% file: latex/Sections/related_work.tex
\section{Related Work}

This work is built on prior research in topic modeling \cite{hofmann1999probabilistic}, supervised \cite{tuarob2015ensemble, peng2019hierarchical, zong2022bgnn} and zero-shot text classification \cite{karmaker2016generalized}, large language models \cite{kojima2022large,sahoo2024systematic,white2023prompt,reynolds2021prompt, jin2024large}, knowledge graph construction \cite{mo2025kggen}, and knowledge graph-enhanced text classification \cite{shi2023chatgraph}. We review the most relevant work in each area below and highlight the remaining challenges that motivate our proposed framework.

\subsection{Topic Modeling and Topic Classification}

\subsubsection{Classical Topic Models:}
The foundation of probabilistic topic discovery was established by PLSA~\cite{hofmann1999probabilistic}. Authors of ~\cite{blei2003latent} later introduced LDA, which added a document-level generative process and became the dominant unsupervised topic model. Subsequent work extended topic modeling through document-relative similarity~\cite{du2015topic}, weakly supervised topic-label mapping~\cite{hingmire2014topic}, and hierarchical Dirichlet processes~\cite{wang2011online}.

\subsubsection{Supervised and Weakly Supervised Topic Classification:}
When labeled data is available, supervised methods can learn topic assignments directly~\cite{tuarob2015ensemble, peng2019hierarchical, zong2022bgnn}. Other work has explored learning from noisy labels~\cite{iwata2009modeling,chen2022neural}, weakly supervised classification~\cite{meng2018weakly}, document annotation~\cite{poursabzi2015speeding}, video categorization~\cite{engels2010automatic}, and domain-specific applications~\cite{hassan2020towards}.

\subsubsection{Zero-Shot Topic Classification:}
A separate line of work considers settings where no labeled examples are available at inference time. In \cite{karmaker2016generalized}, authors explored topic-modeling-based zero-shot methods. Additionally, the authors of \cite{li2016effective} and  \cite{zha2019multilabel} studied dataless classification without annotated training data. Another work \cite{veeranna2016semantic} measured label-document similarity using pretrained word embeddings, and later pursued zero-shot text classification through embedding-based and prompt-based approaches~\cite{rios2018fewshot,xia2018zeroshot,pushp2017train,puri2019zeroshot,chen2021multilabel,gong2021prompt,yin2019benchmarking}. The most directly relevant prior work is that of ~\cite{sarkar2023zeroshot}, in which they formalized zero-shot multi-label topic inference and benchmarked sentence encoders and LLMs. Their results showed that Sentence-BERT was the strongest embedding model among the sentence-encoder baselines they evaluated~\cite{sarkar2023zeroshot,reimers2019sentence}. 
The authors of ~\cite{vannooten2026threshold} further showed that fixed global thresholds perform poorly when similarity distributions vary across models and label sets.

\subsection{Large Language Models and Prompting:}

LLMs have shown strong zero-shot classification ability, assigning labels from natural-language task descriptions without task-specific training~\cite{brown2020language,yin2019benchmarking,chae2025large,vandemoortele2025haystack}. Prior work has also shown that prompt design can substantially affect model behavior and downstream performance~\cite{kojima2022large,sahoo2024systematic,white2023prompt,reynolds2021prompt, jin2024large}. In ~\cite{jang2023negated}, the authors further tested the limits of prompt understanding under negated instructions. 


\subsection{Knowledge Graph Construction from Text:}

Knowledge graph (KG) extraction from unstructured text is a longstanding problem in information extraction and knowledge acquisition~\cite{ji2022survey}. Early rule-based systems such as YAGO~\cite{suchanek2007yago} relied on hard-coded rules. Then OpenIE~\cite{angeli2015leveraging} improved on this using dependency parsing to extract triples (subject, relation, object). However, both of them tend to produce overly specific, inconsistent predicates.
More recent transformer-based extraction pipelines~\cite{qiao2022joint} and external knowledge base approaches such as entity linking to Wikidata or ConceptNet offer better quality but they require fixed relation schemas, domain-specific supervision, or entities present in a pre-built knowledge base. More recently, KGGen \cite{mo2025kggen} introduced an LLM-based framework that produces triples in several stages.

\subsection{Knowledge Graph-Enhanced Text Classification:}

A growing body of work has explored how knowledge graphs can improve text classification. The authors of~\cite{wang2017combining} augmented a CNN-based classifier with concept mappings from an external knowledge base to address limited context in short texts. BERT-KG~\cite{chen2022zeroshot} enriches Sentence-BERT representations with ConceptNet-based knowledge for zero-shot social media classification. Moreover, ChatGraph~\cite{shi2023chatgraph} extracts knowledge graphs from text and uses the resulting graph representations to train an interpretable classifier. Another approach ~\cite{liu2023enhancing} incorporated external knowledge into a hierarchical text classification model through a knowledge-aware encoder and hierarchical label attention. KG-HTC~\cite{zang2025kghtc} retrieves relevant subgraphs from a label-taxonomy knowledge graph and provides them to an LLM as structured context for zero-shot hierarchical text classification. Other related work has explored  KG based data expansion under limited labeled data~\cite{zhang2023towards} and graph-based similarity methods enriched with external knowledge~\cite{shanavas2021knowledge}. Despite these advances, most existing approaches rely on external knowledge bases or supervised training~\cite{chen2022zeroshot,shi2023chatgraph,liu2023enhancing,zang2025kghtc,shanavas2021knowledge}.

\subsection{Self-Consistency Decoding:}
Because LLM outputs are stochastic, a single decoding pass may be unreliable. Wang et al.~\cite{wang2022selfconsistency} proposed self-consistency by sampling multiple outputs at non-zero temperature and aggregating them. Since then, self-consistency decoding has been used for various tasks involving LLM reasoning to decrease the variability of predictions, including mathematical and commonsense reasoning \cite{lin2024just}, program repair \cite{ahmed2023better}, and biomedical natural language inference \cite{liu2024fzi}. Adaptive variants have also been proposed to reduce the computational cost of repeated sampling while preserving accuracy gains \cite{aggarwal2023let, li2024escape}.

\subsection{Differences from Previous Works:}
Considerable progress has been made in zero-shot topic classification, LLM prompting, knowledge graph construction, and knowledge graph based text classification. Nevertheless, these research tracks have largely been pursued independently of one another. Most zero-shot topic classification methods rely on flat document representations, and knowledge graph based classifiers typically require external knowledge bases or supervision during training. As a result, very little 
work has explored the use of document level relational knowledge for LLM based zero-shot multi-label topic classification. In this study, we address this gap by constructing a knowledge graph for each individual document and incorporating the resulting relational structure directly within LLM prompting for zero-shot topic classification. Unlike prior work on knowledge graph-enhanced text classification, our framework operates without relying on external knowledge bases, predefined relation schemas, labeled training data, or task specific fine tuning. Through an exploratory study we present meaningful insights into how document level relational information affects zero-shot multi-label topic classification performance across models of varying scale.

%% file: latex/Sections/problem_statement.tex
\section{Problem Statement}
 
In traditional topic classification, a labeled training dataset and a fixed set of predefined labels are used to train the model in a supervised manner~\cite{alghamdi2015survey}. In real world applications, however, label sets are rarely fixed: they are application dependent and vary across users, making it difficult to anticipate the target labels or collect corresponding labeled data in advance. This motivates a zero shot formulation~\cite{yin2019benchmarking}, where the user specifies the label set only at inference time, without any labeled training data. We define our problem as follows:

\begin{definition}
\label{def:task}
Let a collection of documents $\mathcal{D} = \{d_1, d_2, \ldots, d_n\}$ and a
set of user-defined topics $\mathcal{T}_x = \{t_1, \ldots, t_m\}$ provided at inference time. Each topic $t_j \in \mathcal{T}_x$ is expressed as a word or short phrase and may optionally be associated with a set of descriptive keywords $\mathcal{K}_{t_j} = \{k_1, k_2, \ldots, k_r\}$. The objective is to learn a prediction function


\begin{equation}
f:(D,T_x,K)\rightarrow \mathcal{P}(T_x),
\end{equation}

where $\mathcal{P}(T_x)$ denotes the power set of $T_x$. For every document $d_i$, the model predicts a subset of relevant topics from $\mathcal{P}(T_x)$, where the predicted subset may contain zero, one, or multiple topics. 

\begin{equation}
Y_i=f(d_i,T_x,K)\subseteq T_x,
\end{equation}

without using any labeled training examples for the topics in $T_x$. Here, $Y_i$ denotes the subset of user-defined topics assigned to
document $d_i$.

\end{definition}



The multi-label nature of this task reflects that a document may be associated with several topics at once; for example, a product review may discuss both \textit{Button} and \textit{Screen}. Topics need not appear explicitly by name in the document: a passage discussing \textit{speakers}, \textit{volume}, and \textit{audio quality} may never mention \textit{Sound} directly, yet clearly pertains to it. The optional keyword list $\mathcal{K}_t$ helps disambiguate user intent, since two users may define different topic sets for the same document, but keyword presence or absence is not a reliable indicator on its own: a keyword may appear without its topic being relevant, and a relevant topic may be present without any of its keywords appearing. Keywords therefore serve only as supplementary clues rather than definitive signals.


%% file: latex/Sections/methodology.tex
\section{Method for Zero-shot Topic Inference}
\label{sec:methodoloy}

\begin{figure*}[t]
  \centering
  \includegraphics[width=\textwidth, trim=80 95 40 100, clip]{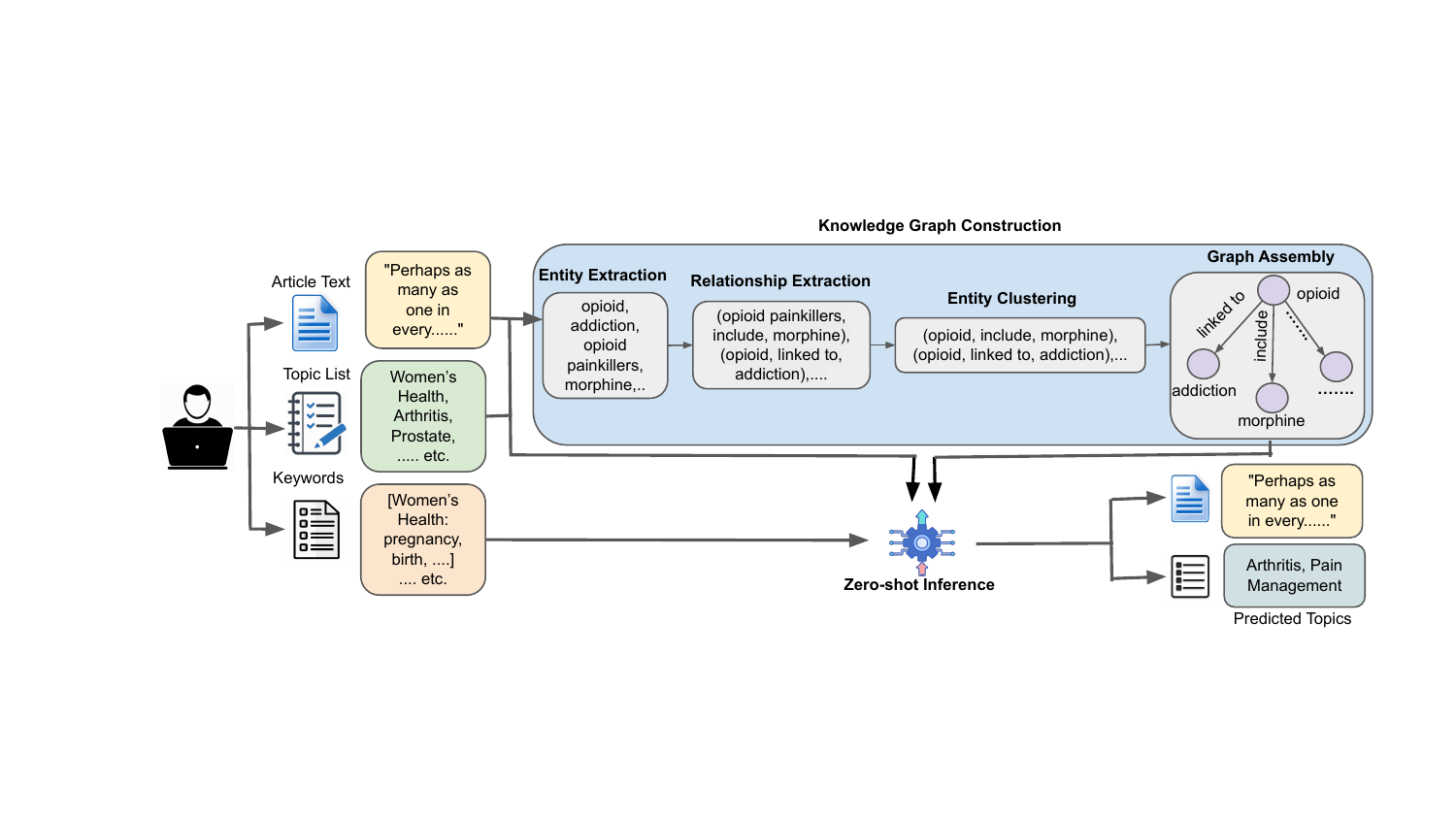}
  \caption{Overall architecture of the proposed KG-enhanced zero-shot multi-label topic inference system. Each article first passes through a four-stage KG construction pipeline to produce a per-article knowledge graph. The graph, corresponding article, topic list and an optional keyword list are then fed into four graph-enhanced classification variants: AG (Article + Graph), AKG (AG + Keywords), AGS (AG + self-consistency), and AKGS (AKG + self-consistency).}
  \label{fig:architecture}
\end{figure*}

In this section, we discuss the knowledge graph-enhanced zero-shot topic inference approach studied in this paper. The end-to-end inference process is shown in Fig.~\ref{fig:architecture}.

\begin{enumerate}
    \item The end user provides the inputs, i.e., the article text, user defined candidate topics, and optionally a set of descriptive keywords for each topic.
    
    \item For each article, a document level knowledge graph is constructed using an LLM driven extraction pipeline inspired by KGGen~\cite{mo2025kggen}. This construction proceeds in four stages  (described in detail in Section~\ref{subsec:kg_construction}): entity extraction, relation extraction, entity clustering, and graph assembly. The same LLM used for this extraction step later serves as the backbone for topic classification.
    
    \item The resulting knowledge graph, represented as a set of subject predicate object triples, is cached to disk and reused across all downstream classification experiments for that LLM, avoiding redundant regeneration.
    
    \item At inference time, the article text, the candidate topics (with optional keywords), and, the corresponding knowledge graph are jointly provided to the LLM as input.
    
    \item The LLM performs zero-shot multi-label classification directly from this input, without any task specific fine tuning or labeled training examples, producing a predicted subset of relevant topics for the article. We evaluate eight variations of this inference framework in total, four that incorporate the knowledge graph and four that do not, allowing us to directly assess the contribution of graph augmentation to classification performance.
    
    \item The output of the zero-shot topic inference framework is the set of inferred topic(s) for each article, which may contain zero, one, or multiple topics.
\end{enumerate}

To isolate and quantify the effect of the graph, we additionally perform the same inference process without providing the KG as input, using only the article text, candidate topics, and optional keywords. Comparing the graph-enhanced and graph-free variants under identical conditions helps us to directly measure the extent to which the knowledge graph contributes to zero-shot multi-label topic classification performance.

\subsection{Knowledge Graph Construction}
\label{subsec:kg_construction}
For each article, a knowledge graph is constructed from the article text following the extraction methodology of the KGGen framework~\cite{mo2025kggen}, a plain text to knowledge graph construction pipeline that uses LLMs to extract structured subject-predicate-object triples from unstructured text. KGGen~\cite{mo2025kggen} introduces a multi-stage design comprising entity extraction, relation extraction, and a subsequent clustering stage that merges synonymous nodes and edges across an entire corpus. We adapt this framework to apply independently to each article in the dataset. Each of the fifteen LLMs evaluated in this study serves as the backbone model for its own extraction step, that is, the same LLM used to construct a document's knowledge graph is also the one subsequently used for downstream topic classification. Each knowledge graph is constructed once per LLM and stored in JSON format, ensuring that the same graph is reused consistently across every classification experiment conducted for that LLM.

\begin{table}[t]
\caption{Description of Eight classification variants of the inference pipeline}
\label{tab:Variants}
\centering
\renewcommand{\arraystretch}{1.2}
\begin{tabularx}{\linewidth}{>{\raggedright\arraybackslash\bfseries}p{1.8cm}|>{\justifying\arraybackslash}X}
\hline
\textbf{Name of the Method} & \textbf{Description} \\
\hline
AO (Article only) &
The model receives just the article text and the topic names without any additional information. The model has to completely depend on the knowledge it has about what each topic name means and the information provided by the article.
\\
\hline
AK (Article + Keywords) &
The model is also provided with a per-topic keyword list. Every dataset we test on~\cite{sarkar2023zeroshot} comes with a corresponding keyword file that contains keywords which have strong association with each topic, arranged as: \texttt{topic (keywords: k1, k2, k3, \ldots)}. The model is asked to make use of these keywords to understand what the topic means. The keywords guide the model to connect the article content to the topic. \\
\hline
AOS (Article Only + Self-Consistency Decoding) &
Self-consistency~\cite{wang2022selfconsistency} addresses the stochastic nature of single-pass LLM generation: the same prompt may produce different outputs on different runs. In AOS, the Article Only prompt is run $N=5$ times and topics that appear in at least three runs are retained in the final prediction. In this way, self-consistency serves as noise reduction. Topics which repeat consistently are more likely to be valid, whereas those which occur in a single iteration are treated as noise. \\
\hline
AKS (Article + Keywords + Self-Consistency Decoding) &
In AKS, keyword-based classification is used together with self-consistency decoding based on the same majority voting mechanism as in AOS. The input to the system includes the article, topic list and the keywords. \\
\hline
AG (Article + Graph) &
This is AO variant augmented with the graph. The model receives the article, the topic list, and the generated knowledge graph associated with that article. The graph offers structured context regarding the entities and relations between them.
\\
\hline
AKG (Article + Keywords + Graph) &
This is AK augmented with the graph. The model receives article, topic list, the keyword list associated with the topics and the knowledge graph generated. \\
\hline
AGS (Article + Graph + self-consistency) &
It is an extension of AG in which the self-consistency decoding technique is used.  \\
\hline
AKGS (Article + Keywords + Graph + self-consistency) &
This is the most extensive setting. The model receives the article, topic list, the per-topic keyword list, and the graph. The predictions are aggregated using self-consistency decoding, similar to AGS. \\
\hline
\end{tabularx}
\end{table}

\subsection{Zero-shot Inference Pipeline}
\label{subsec:inference_pipeline}

Using combinations of the article text, candidate topics, optional keywords, and the constructed graph, we evaluate eight variants of the zero-shot inference pipeline. These variants are designed to isolate the individual and combined contributions of keyword guidance, knowledge graph augmentation, and self-consistency decoding. The variants fall into two broad groups: four graph free variants (AO, AK, AOS, AKS), and four graph enhanced variants (AG, AKG, AGS, AKGS). Table~\ref{tab:Variants} summarizes all eight variants along with their input configurations.

%% file: latex/Sections/experimental_setup.tex
\section{Experimental Setup}

In this section, we describe the datasets, models, evaluation metrics, and implementation details that were used in our experiments.

\begin{table}[h]
\caption{Statistics of the Datasets}
\label{tab:datasets}
\centering
\footnotesize
\setlength{\tabcolsep}{6pt}
\renewcommand{\arraystretch}{1.05}
\begin{tabular}{lrrrr}
\toprule
\textbf{Dataset} & 
\textbf{Articles} & 
\makecell{\textbf{Avg.}\\\textbf{Length}} & 
\textbf{Topics} & 
\makecell{\textbf{Topics/}\\\textbf{Article}} \\
\midrule
Medical          & 2066 & 693 & 18 & 1.128 \\
News             & 8940 & 589 & 12 & 0.805 \\
Cellular phone   & 587  & 16  & 23 & 1.058 \\
Digital camera 1 & 642  & 18  & 24 & 1.069 \\
Digital camera 2 & 380  & 17  & 20 & 1.039 \\
DVD player       & 839  & 15  & 23 & 0.781 \\
Mp3 player       & 1811 & 17  & 21 & 0.956 \\
SemEval          & 3259 & 16  & 11 & 2.415 \\
\bottomrule
\end{tabular}
\end{table}

\begin{table}[t]
\caption{List of Evaluated large language models}
\label{tab:llms}
\centering
\setlength{\tabcolsep}{6pt}
\renewcommand{\arraystretch}{1.05}
\begin{tabular}{llr}
\toprule
\textbf{Category} & \textbf{Model} & \textbf{Parameters} \\
\midrule
\multirow{5}{*}{\shortstack{Small\\(Less than 5B)}}
& Gemma 3n-E4B & E4B \\
& Llama 3.2-3B & 3B \\
& Qwen2.5-3B & 3B \\
& Ministral-3B & 3B \\
& DeepSeek-R1-Distill-Qwen-1.5B & 1.5B \\
\midrule
\multirow{5}{*}{\shortstack{Medium\\ (From 4B to 20B)}}
& GPT-OSS-20B & 20B \\
& Mixtral-8$\times$7B & 8$\times$7B \\
& Gemma 2-9B & 9B \\
& Llama 3.1-8B & 8B \\
& Qwen2.5-7B & 7B \\
\midrule
\multirow{5}{*}{\shortstack{Large \\(More than 20B)}}
& Llama 3.3-70B & 70B \\
& Qwen2.5-72B & 72B \\
& Qwen3-32B & 32B \\
& Gemma 3-27B & 27B \\
& GPT-4o & Not disclosed \\
\bottomrule
\end{tabular}
\end{table}

\subsection{Datasets}

Our proposed framework was evaluated on eight multi-label topic classification datasets, including five product review datasets (Cellular phone, Digital camera 1, Digital camera 2, DVD player, and Mp3 player) \cite{sarkar2023zeroshot}, two large datasets (Medical and News) \cite{sarkar2023zeroshot}, and the English subset of the SemEval 2018 dataset \cite{mohammad2018semeval}, a social media dataset consisting of tweets annotated for multi-label emotion classification. These datasets were deliberately chosen to span diverse domains, document lengths, label cardinalities, and writing styles, ranging from short, informal, product reviews and tweets to long form, formal medical and news text, in order to rigorously assess the generalizability and robustness of our framework across heterogeneous data types. An overview of the datasets is provided in Table \ref{tab:datasets}.

In zero-shot learning, the end user can provide auxiliary information about each topic in the form of keywords. All seven datasets from \cite{sarkar2023zeroshot} contain keyword files associated with each dataset, where the keywords serve as auxiliary information representing words related to each topic in that dataset. The eighth dataset, SemEval 2018 Task~1 \cite{mohammad2018semeval}, a multi-label emotion classification benchmark, did not have an associated keyword file. We therefore generated per-topic keywords following the approach in \cite{sarkar2022concept}, adopting the same keyword format used for the other seven datasets.

\subsection{Large Language Models}

We evaluate fifteen LLMs spanning a broad range of model families and parameter sizes (Table \ref{tab:llms}). For each model, the same LLM is used for both knowledge graph construction and zero-shot topic classification. This allows a consistent end-to-end evaluation of the complete pipeline.

\begin{table*}[t]
\centering
\caption{Prompt design details for the knowledge graph construction pipeline: triple extraction and entity cluster validation.}
\label{tab:kg_prompt_design}
\renewcommand{\arraystretch}{1.15}
\begin{tabularx}{\textwidth}{l|X}
\hline
\multicolumn{2}{c}{\textbf{Step 1: Entity Extraction}} \\
\hline
\textbf{Prompt} & "Extract key entities from the given text. Extracted entities are nouns, verbs, or adjectives, particularly regarding sentiment. This is for an extraction task, please be thorough, accurate, and faithful to the reference to the reference text.

Return ONLY a valid JSON list format: ["entity1", "entity2", "entity3"]" \\
\hline
\textbf{Input} & Article: "Perhaps as many as one in every 5 American adults will get a prescription for a painkiller this year, and many more will buy over-the-counter medicines without a prescription. These drugs can do wonders getting rid of pain can seem like a miracle but sometimes......" \\
\hline
\textbf{Output} & A JSON array of entities: \texttt{["opioid","addiction","morphine",.....]} \\
\hline
\multicolumn{2}{c}{\textbf{Step 2: Subject-Predicate-Object (SPO) Triple Extraction}} \\
\hline
\textbf{Prompt} & "Extract subject-predicate-object triples from the assistant message. A predicate (1-3 words) defines the relationship between the subject and object. Relationship may be fact or sentiment based on assistant's message. Subject and object are entities. Entities are provided as input, though you may not need all of them. This is for an extraction task, please be thorough, accurate, and faithful to the reference text.

Return ONLY valid JSON format: [["subject1", "predicate1", "object1"], ["subject2", "predicate2", "object2"],....]" \\
\hline
\textbf{Input} & Article: "Perhaps as many as one in every 5 American adults will get a prescription for a painkiller this year, and many more will buy over-the-counter medicines without a prescription. These drugs can do wonders getting rid of pain can seem like a miracle but sometimes......" \newline

A JSON array of entities extracted from the previous step:
["opioid","addiction","morphine",.....]
 \\
\hline
\textbf{Output} & A JSON array of triples:
[["opioid painkillers","include","morphine"], ["opioid","linked to","addiction"],....] \\
\hline
\multicolumn{2}{c}{\textbf{Step 3: Entity Cluster Validation}} \\
\hline
\textbf{Prompt} & "Verify if these entities belong in the same cluster. A cluster should contain entities that are the same in meaning, with different: tenses, plural forms, stem forms, upper/lower cases Or entities with close semantic meanings.

\par\medskip
Return ONLY valid JSON format: ["entity1", "entity2", "entity3"]. 
Return only entities you are confident belong together.
    If not confident, return empty list [].
    " \\
\hline
\textbf{Input} & A JSON array of triples: [["opioid painkillers","include","morphine"], ["opioid","linked to","addiction"],...] \\
\hline
\textbf{Output} & A JSON array of triples after clustering the similar entities: [["opioid","include","morphine"], ["opioid","linked to","addiction"],...] \\
\hline
\end{tabularx}
\end{table*}

\begin{table*}[t]
\centering
\caption{Prompt design details for the inference stage used for topic classification. (AKG: Article + Graph + Keywords)}
\label{tab:inference_prompt_design}
\renewcommand{\arraystretch}{1.15}
\begin{tabularx}{\textwidth}{l|X}
\hline
\multicolumn{2}{c}{\textbf{Prompt Design}} \\
\hline
\textbf{Prompt} & "Given an article text, its knowledge graph, and a list of possible topics with their associated keywords, determine which topics (can be multiple) are most relevant to this article. Use the keywords to better understand what each topic represents. Match the article content and knowledge graph entities/relations against these keywords. 
\par\medskip
Return only topic names that are actually present in the available topics list.
If no topics meet the strong-evidence criteria, return `None`." \\
\hline
\textbf{Input} & 1. Article: "Perhaps as many as one in every 5 American adults will get a prescription for a painkiller this year, and many more will buy over-the-counter medicines without a prescription. These drugs can do wonders getting rid of pain can seem like a miracle but sometimes......"
\newline
2. The knowledge graph constructed from this specific article
\newline
3. Topic list: [“Arthritis”, “Heart Health”, “Pain Management”,....]
\newline
4. Keyword list: [["Keyword":["Arthritis","pain","knee",....]],....]
 \\
\hline
\textbf{Output} & Topics: [“Arthritis”, “Pain Management”,....] \\
\hline
\end{tabularx}
\end{table*}

\subsection{Evaluation Metrics}

We used three widely used metrics: Precision, Recall, and F1-score \cite{sokolova2009systematic} to evaluate our framework. First, we take inferred topics and compare them against the ground-truth topics to derive true positive, false positive, and false negative counts. The total TP, FP, and FN values are calculated for all the articles. We calculated micro-average of the F1-score by aggregating the global counts of TP, FP, and FN across all instances.


\subsection{Implementation Details}
All experiments were implemented in Python 3.11 using DSPy for LLM management and structured prompting. Knowledge graphs are generated once per LLM per dataset and cached to disk as JSON files, avoiding redundant regeneration across experiments. API calls are made sequentially; failed calls yield an empty prediction, counted as all false negatives during evaluation. The self-consistency variant issues five times as many API calls per experiment. Temperature is set to 0.3 for single-pass variants and 0.5 for self-consistency decoding.

\subsection{Prompt Design}
The performance of the LLM-based knowledge graph construction and topic inference framework is mainly determined by task-specific prompt design. In our research, we design prompts in order to ensure the consistency of the outputs of the LLMs. This includes task definition, inputs, outputs, and constraints. Table~\ref{tab:kg_prompt_design} is an example of the prompts designed for graph construction and Table~\ref{tab:inference_prompt_design} shows the prompts applied in the inference process for topic classification.

%% file: latex/Sections/results_and_analysis.tex
\section{Results and Analysis}
\label{sec:results}

\begin{table*}[t]
\caption{F1-scores for the eight inference methods on smaller models. Graph-free methods (AO, AK, AOS, AKS) and graph-enhanced methods (AG, AKG, AGS, AKGS) are shown side by side. AO~=~Article Only; AK~=~Article + Keywords; AOS~=~AO + Self-Consistency; AKS~=~AK + Self-Consistency; AG~=~Article + Graph; AKG~=~AG + Keywords; AGS~=~AG + Self-Consistency; AKGS~=~AKG + Self-Consistency. DS-R1~1.5B~=~DeepSeek-R1-Distill-Qwen-1.5B. Bold values exceed the baseline~\cite{sarkar2023zeroshot} (Base.\ column).} 
\renewcommand{\arraystretch}{1.1}
\centering
\scriptsize
\setlength{\tabcolsep}{2pt}
\resizebox{\textwidth}{!}{%
\begin{tabular}{|l|c|l|ccccc|l|ccccc|}
\hline

\multicolumn{1}{|l|}{\textbf{}}
& \multicolumn{1}{c|}{\textbf{}}
& \multicolumn{1}{l|}{\textbf{}}
& \multicolumn{5}{c|}{\textbf{Graph-free}}
& \multicolumn{1}{l|}{\textbf{}}
& \multicolumn{5}{c|}{\textbf{Graph-enhanced}} \\

\cline{4-8} \cline{10-14}

\textbf{Dataset}
& \textbf{Base.}
& \textbf{Method}
& \shortstack{\textbf{Gemma} \\ \textbf{3n-E4B}}
& \shortstack{\textbf{LLaMA} \\ \textbf{3.2-3B}}
& \shortstack{\textbf{Qwen} \\ \textbf{2.5-3B}}
& \shortstack{\textbf{Ministral} \\ \textbf{3B}}
& \shortstack{\textbf{DS-R1} \\ \textbf{1.5B}}
& \textbf{Method}
& \shortstack{\textbf{Gemma} \\ \textbf{3n-E4B}}
& \shortstack{\textbf{LLaMA} \\ \textbf{3.2-3B}}
& \shortstack{\textbf{Qwen} \\ \textbf{2.5-3B}}
& \shortstack{\textbf{Ministral} \\ \textbf{3B}}
& \shortstack{\textbf{DS-R1} \\ \textbf{1.5B}} \\[4pt]

\hline


\multirow{4}{*}{Medical}
& \multirow{4}{*}{0.606} & \textsc{ao}   & 0.574 & 0.548 & 0.539 & 0.508 & 0.488 & \textsc{ag}   & 0.586 & 0.575 & 0.551 & 0.539 & 0.497 \\
&       & \textsc{ak}   & 0.585 & 0.570 & 0.557 & 0.539 & 0.497 & \textsc{akg}  & 0.592 & 0.580 & 0.579 & 0.564 & 0.533 \\
&       & \textsc{aos}  & 0.551 & 0.530 & 0.542 & 0.496 & 0.484 & \textsc{ags}  & 0.559 & 0.558 & 0.549 & 0.536 & 0.493 \\
&       & \textsc{aks}  & 0.544 & 0.537 & 0.525 & 0.511 & 0.472 & \textsc{akgs} & 0.577 & 0.571 & 0.558 & 0.545 & 0.495 \\

\hline


\multirow{4}{*}{News}
& \multirow{4}{*}{0.521} & \textsc{ao}   & 0.503 & 0.498 & 0.477 & 0.463 & 0.422 & \textsc{ag}   & 0.514 & 0.498 & 0.483 & 0.471 & 0.441 \\
&       & \textsc{ak}   & 0.505 & 0.507 & 0.487 & 0.475 & 0.433 & \textsc{akg}  & 0.510 & 0.515 & 0.491 & 0.486 & 0.457 \\
&       & \textsc{aos}  & 0.499 & 0.493 & 0.469 & 0.460 & 0.421 & \textsc{ags}  & 0.500 & 0.495 & 0.480 & 0.468 & 0.438 \\
&       & \textsc{aks}  & 0.481 & 0.484 & 0.459 & 0.443 & 0.406 & \textsc{akgs} & 0.502 & 0.507 & 0.490 & 0.485 & 0.447 \\

\hline


\multirow{4}{*}{Cell. phone}
& \multirow{4}{*}{0.576} & \textsc{ao}   & 0.519 & 0.367          & 0.463 & 0.459          & 0.415 & \textsc{ag}   & 0.528 & 0.381          & 0.481 & 0.469          & 0.413 \\
&       & \textsc{ak}   & 0.554 & 0.345          & 0.495 & \textbf{0.594} & 0.429 & \textsc{akg}  & 0.555 & 0.351          & 0.499 & \textbf{0.600} & 0.431 \\
&       & \textsc{aos}  & 0.507          & 0.317 & 0.482 & 0.399          & 0.409 & \textsc{ags}  & 0.508          & 0.329 & 0.478 & 0.397          & 0.408 \\
&       & \textsc{aks}  & 0.494          & 0.314          & 0.464 & 0.510          & 0.398 & \textsc{akgs} & 0.513          & 0.433          & 0.498 & 0.533 & 0.469 \\

\hline


\multirow{4}{*}{Digital cam. 1}
& \multirow{4}{*}{0.641} & \textsc{ao}   & 0.488 & 0.435 & 0.463 & 0.486          & 0.395 & \textsc{ag}   & 0.501 & 0.443 & 0.461 & 0.484          & 0.471 \\
&       & \textsc{ak}   & 0.510 & 0.411 & 0.478 & 0.514 & 0.401 & \textsc{akg}  & 0.528 & 0.415 & 0.479 & 0.517 & 0.407 \\
&       & \textsc{aos}  & 0.482          & 0.477 & 0.458 & 0.372          & 0.388 & \textsc{ags}  & 0.482          & 0.473 & 0.458 & 0.469          & 0.387 \\
&       & \textsc{aks}  & 0.478          & 0.269 & 0.447 & 0.415          & 0.379 & \textsc{akgs} & 0.498          & 0.393 & 0.468 & 0.437          & 0.398 \\

\hline


\multirow{4}{*}{Digital cam. 2}
& \multirow{4}{*}{0.562} & \textsc{ao}   & 0.552 & 0.458 & 0.548 & 0.417 & 0.374 & \textsc{ag}   & 0.553 & 0.455 & 0.549 & 0.416 & 0.372 \\
&       & \textsc{ak}   & \textbf{0.580} & 0.466 & \textbf{0.563} & 0.480 & 0.387 & \textsc{akg}  & \textbf{0.594} & 0.469 & \textbf{0.567} & 0.485 & 0.389 \\
&       & \textsc{aos}  & \textbf{0.568} & 0.538 & 0.549 & 0.382 & 0.367 & \textsc{ags}  & \textbf{0.576} & 0.535 & 0.546 & 0.382 & 0.368 \\
&       & \textsc{aks}  & 0.547 & 0.439 & 0.537 & 0.454 & 0.357 & \textsc{akgs} & \textbf{0.570} & 0.460 & 0.556 & 0.478 & 0.381 \\

\hline


\multirow{4}{*}{DVD player}
& \multirow{4}{*}{0.533} & \textsc{ao}   & 0.469 & 0.265 & 0.431 & 0.397          & 0.339 & \textsc{ag}   & 0.476 & 0.367 & 0.431 & 0.398          & 0.341 \\
&       & \textsc{ak}   & 0.496 & 0.323 & 0.446 & 0.521 & 0.352 & \textsc{akg}  & 0.499 & 0.425 & 0.449 & \textbf{0.546} & 0.358 \\
&       & \textsc{aos}  & 0.472 & 0.288 & 0.427 & 0.351          & 0.340 & \textsc{ags}  & 0.472 & 0.386 & 0.428 & 0.347          & 0.337 \\
&       & \textsc{aks}  & 0.465 & 0.337 & 0.414 & 0.406          & 0.330 & \textsc{akgs} & 0.488 & 0.357 & 0.438 & 0.425          & 0.349 \\

\hline


\multirow{4}{*}{Mp3 player}
& \multirow{4}{*}{0.571} & \textsc{ao}   & 0.493 & 0.355 & 0.445 & 0.507          & 0.378 & \textsc{ag}   & 0.489 & 0.361 & 0.443 & 0.507          & 0.374 \\
&       & \textsc{ak}   & 0.512 & 0.394 & 0.457 & \textbf{0.641} & 0.385 & \textsc{akg}  & 0.517 & 0.397 & 0.461 & \textbf{0.645} & 0.388 \\
&       & \textsc{aos}  & 0.487 & 0.347 & 0.440 & 0.406          & 0.372 & \textsc{ags}  & 0.485 & 0.347 & 0.440 & 0.403          & 0.370 \\
&       & \textsc{aks}  & 0.480 & 0.375 & 0.429 & 0.522 & 0.358 & \textsc{akgs} & 0.501 & 0.398 & 0.450 & \textbf{0.577} & 0.381 \\

\hline


\multirow{4}{*}{SemEval}
& \multirow{4}{*}{0.550} & \textsc{ao}   & 0.521 & 0.446 & 0.470 & 0.456 & 0.404 & \textsc{ag}   & \textbf{0.555} & 0.445 & 0.468 & 0.456 & 0.401 \\
&       & \textsc{ak}   & 0.545 & 0.518 & 0.484 & 0.476 & 0.412 & \textsc{akg}  & 0.548 & 0.523 & 0.486 & 0.480 & 0.417 \\
&       & \textsc{aos}  & 0.518 & 0.474 & 0.466 & 0.475 & 0.389 & \textsc{ags}  & 0.519 & 0.471 & 0.465 & 0.472 & 0.397 \\
&       & \textsc{aks}  & 0.515 & 0.399 & 0.452 & 0.439 & 0.388 & \textsc{akgs} & 0.535 & 0.421 & 0.475 & 0.462 & 0.409 \\

\hline
\end{tabular}
}
\label{tab:small-results}
\end{table*}

\begin{table*}[t]
\caption{F1-scores for the eight inference methods on medium-size models. Graph-free methods (AO, AK, AOS, AKS) and graph-enhanced methods (AG, AKG, AGS, AKGS) are shown side by side. AO~=~Article Only; AK~=~Article + Keywords; AOS~=~AO + Self-Consistency; AKS~=~AK + Self-Consistency; AG~=~Article + Graph; AKG~=~AG + Keywords; AGS~=~AG + Self-Consistency; AKGS~=~AKG + Self-Consistency. Bold values exceed the baseline~\cite{sarkar2023zeroshot} (Base.\ column).}
\renewcommand{\arraystretch}{1.1}
\centering
\scriptsize
\setlength{\tabcolsep}{2pt}
\resizebox{\textwidth}{!}{%
\begin{tabular}{|l|c|l|ccccc|l|ccccc|}
\hline

\multicolumn{1}{|l|}{\textbf{}}
& \multicolumn{1}{c|}{\textbf{}}
& \multicolumn{1}{l|}{\textbf{}}
& \multicolumn{5}{c|}{\textbf{Graph-free}}
& \multicolumn{1}{l|}{\textbf{}}
& \multicolumn{5}{c|}{\textbf{Graph-enhanced}} \\

\cline{4-8} \cline{10-14}

\textbf{Dataset}
& \textbf{Base.}
& \textbf{Method}
& \shortstack{\textbf{GPT-OSS} \\ \textbf{20B}}
& \shortstack{\textbf{Mixtral} \\ \textbf{8x7B}}
& \shortstack{\textbf{Gemma} \\ \textbf{2-9B}}
& \shortstack{\textbf{LLaMA} \\ \textbf{3.1-8B}}
& \shortstack{\textbf{Qwen} \\ \textbf{2.5-7B}}
& \textbf{Method}
& \shortstack{\textbf{GPT-OSS} \\ \textbf{20B}}
& \shortstack{\textbf{Mixtral} \\ \textbf{8x7B}}
& \shortstack{\textbf{Gemma} \\ \textbf{2-9B}}
& \shortstack{\textbf{LLaMA} \\ \textbf{3.1-8B}}
& \shortstack{\textbf{Qwen} \\ \textbf{2.5-7B}} \\[4pt]

\hline


\multirow{4}{*}{Medical}
& \multirow{4}{*}{0.606} & \textsc{ao}   & \textbf{0.636} & 0.606 & \textbf{0.634} & \textbf{0.676} & \textbf{0.698} & \textsc{ag}   & \textbf{0.644} & \textbf{0.614} & \textbf{0.631} & \textbf{0.657} & \textbf{0.688} \\
&       & \textsc{ak}   & \textbf{0.671} & \textbf{0.657} & \textbf{0.671} & 0.519          & \textbf{0.671} & \textsc{akg}  & \textbf{0.676} & \textbf{0.660} & \textbf{0.673} & 0.496          & \textbf{0.714} \\
&       & \textsc{aos}  & \textbf{0.624} & \textbf{0.612} & \textbf{0.631} & \textbf{0.663} & \textbf{0.702} & \textsc{ags}  & \textbf{0.621} & \textbf{0.610} & \textbf{0.628} & \textbf{0.633} & \textbf{0.697} \\
&       & \textsc{aks}  & \textbf{0.631} & \textbf{0.614} & \textbf{0.626} & 0.436          & 0.568          & \textsc{akgs} & \textbf{0.651} & \textbf{0.638} & \textbf{0.649} & 0.424          & \textbf{0.703} \\

\hline


\multirow{4}{*}{News}
& \multirow{4}{*}{0.521} & \textsc{ao}   & \textbf{0.655} & \textbf{0.635} & \textbf{0.649} & \textbf{0.622} & \textbf{0.609} & \textsc{ag}   & \textbf{0.672} & \textbf{0.630} & \textbf{0.648} & \textbf{0.623} & \textbf{0.611} \\
&       & \textsc{ak}   & \textbf{0.693} & \textbf{0.667} & \textbf{0.685} & \textbf{0.660} & \textbf{0.636} & \textsc{akg}  & \textbf{0.698} & \textbf{0.672} & \textbf{0.690} & \textbf{0.661} & \textbf{0.647} \\
&       & \textsc{aos}  & \textbf{0.657} & \textbf{0.626} & \textbf{0.644} & \textbf{0.623} & \textbf{0.588} & \textsc{ags}  & \textbf{0.654} & \textbf{0.626} & \textbf{0.645} & \textbf{0.619} & \textbf{0.607} \\
&       & \textsc{aks}  & \textbf{0.660} & \textbf{0.632} & \textbf{0.641} & \textbf{0.619} & \textbf{0.605} & \textsc{akgs} & \textbf{0.679} & \textbf{0.651} & \textbf{0.666} & \textbf{0.639} & \textbf{0.626} \\

\hline


\multirow{4}{*}{Cell. phone}
& \multirow{4}{*}{0.576} & \textsc{ao}   & \textbf{0.677} & \textbf{0.579} & \textbf{0.614} & \textbf{0.587} & \textbf{0.629} & \textsc{ag}   & \textbf{0.684} & \textbf{0.592} & \textbf{0.611} & \textbf{0.587} & \textbf{0.631} \\
&       & \textsc{ak}   & \textbf{0.752} & \textbf{0.615} & \textbf{0.648} & \textbf{0.619} & \textbf{0.637} & \textsc{akg}  & \textbf{0.753} & \textbf{0.637} & \textbf{0.653} & \textbf{0.625} & \textbf{0.644} \\
&       & \textsc{aos}  & \textbf{0.695} & \textbf{0.589} & \textbf{0.607} & \textbf{0.584} & \textbf{0.686} & \textsc{ags}  & \textbf{0.692} & \textbf{0.578} & \textbf{0.608} & \textbf{0.583} & \textbf{0.670} \\
&       & \textsc{aks}  & \textbf{0.680} & \textbf{0.588} & \textbf{0.605} & \textbf{0.582} & \textbf{0.666} & \textsc{akgs} & \textbf{0.703} & \textbf{0.611} & \textbf{0.629} & \textbf{0.603} & \textbf{0.689} \\

\hline


\multirow{4}{*}{Digital cam. 1}
& \multirow{4}{*}{0.641} & \textsc{ao}   & \textbf{0.670} & 0.571 & 0.596 & \textbf{0.677} & \textbf{0.674} & \textsc{ag}   & \textbf{0.675} & 0.568 & 0.597 & \textbf{0.679} & \textbf{0.674} \\
&       & \textsc{ak}   & \textbf{0.724} & 0.607 & 0.635 & \textbf{0.696} & \textbf{0.712} & \textsc{akg}  & \textbf{0.729} & 0.611 & 0.639 & \textbf{0.701} & \textbf{0.717} \\
&       & \textsc{aos}  & \textbf{0.678} & 0.565 & 0.593 & \textbf{0.649} & \textbf{0.644} & \textsc{ags}  & \textbf{0.679} & 0.564 & 0.594 & \textbf{0.649} & \textbf{0.646} \\
&       & \textsc{aks}  & 0.577 & 0.564 & 0.596 & \textbf{0.650} & \textbf{0.665} & \textsc{akgs} & \textbf{0.651} & 0.587 & 0.615 & \textbf{0.671} & \textbf{0.686} \\

\hline


\multirow{4}{*}{Digital cam. 2}
& \multirow{4}{*}{0.562} & \textsc{ao}   & \textbf{0.647} & \textbf{0.617} & \textbf{0.584}          & \textbf{0.581}          & \textbf{0.613} & \textsc{ag}   & \textbf{0.653} & \textbf{0.626} & \textbf{0.582}          & \textbf{0.582}          & 0.555          \\
&       & \textsc{ak}   & \textbf{0.683} & \textbf{0.653} & \textbf{0.601} & \textbf{0.614} & \textbf{0.649} & \textsc{akg}  & \textbf{0.684} & \textbf{0.658} & \textbf{0.624} & \textbf{0.619} & \textbf{0.615} \\
&       & \textsc{aos}  & \textbf{0.660} & \textbf{0.621} & \textbf{0.578}          & \textbf{0.567} & \textbf{0.624} & \textsc{ags}  & \textbf{0.656} & \textbf{0.623} & \textbf{0.579}          & \textbf{0.639} & \textbf{0.639} \\
&       & \textsc{aks}  & \textbf{0.571}          & \textbf{0.614} & \textbf{0.575}          & \textbf{0.641} & 0.482          & \textsc{akgs} & \textbf{0.590}          & \textbf{0.635} & \textbf{0.600}          & \textbf{0.661} & \textbf{0.654} \\

\hline


\multirow{4}{*}{DVD player}
& \multirow{4}{*}{0.533} & \textsc{ao}   & \textbf{0.570} & 0.505 & 0.513 & 0.502 & \textbf{0.564} & \textsc{ag}   & \textbf{0.581} & 0.515 & 0.514 & 0.501 & \textbf{0.561} \\
&       & \textsc{ak}   & \textbf{0.655} & \textbf{0.534} & \textbf{0.550} & \textbf{0.605} & \textbf{0.570} & \textsc{akg}  & \textbf{0.659} & \textbf{0.541} & \textbf{0.556} & \textbf{0.606} & \textbf{0.571} \\
&       & \textsc{aos}  & \textbf{0.610} & 0.498          & 0.510 & \textbf{0.534} & \textbf{0.538} & \textsc{ags}  & \textbf{0.611} & 0.503 & 0.511 & 0.531 & \textbf{0.535} \\
&       & \textsc{aks}  & \textbf{0.595} & 0.501          & 0.512 & \textbf{0.605} & 0.532 & \textsc{akgs} & \textbf{0.618} & 0.520 & 0.532 & \textbf{0.624} & \textbf{0.552} \\

\hline


\multirow{4}{*}{Mp3 player}
& \multirow{4}{*}{0.571} & \textsc{ao}   & \textbf{0.645} & 0.516 & 0.536 & 0.495          & \textbf{0.589} & \textsc{ag}   & \textbf{0.657} & 0.528 & 0.537 & 0.495          & \textbf{0.586} \\
&       & \textsc{ak}   & \textbf{0.698} & 0.557 & \textbf{0.573} & \textbf{0.620} & \textbf{0.641} & \textsc{akg}  & \textbf{0.752} & 0.562 & \textbf{0.579} & \textbf{0.624} & \textbf{0.643} \\
&       & \textsc{aos}  & \textbf{0.662} & 0.526 & 0.536 & \textbf{0.606} & \textbf{0.610} & \textsc{ags}  & \textbf{0.660} & 0.524 & 0.534 & \textbf{0.606} & \textbf{0.612} \\
&       & \textsc{aks}  & \textbf{0.676} & 0.530          & 0.536 & \textbf{0.627} & \textbf{0.647} & \textsc{akgs} & \textbf{0.729} & 0.542 & 0.555 & \textbf{0.648} & \textbf{0.666} \\

\hline


\multirow{4}{*}{SemEval}
& \multirow{4}{*}{0.550} & \textsc{ao}   & \textbf{0.604} & \textbf{0.559} & \textbf{0.576} & \textbf{0.564} & \textbf{0.578} & \textsc{ag}   & \textbf{0.615} & \textbf{0.558} & \textbf{0.573} & \textbf{0.561} & \textbf{0.579} \\
&       & \textsc{ak}   & \textbf{0.617} & \textbf{0.589} & \textbf{0.613} & \textbf{0.573} & \textbf{0.591} & \textsc{akg}  & \textbf{0.622} & \textbf{0.593} & \textbf{0.615} & \textbf{0.579} & \textbf{0.595} \\
&       & \textsc{aos}  & \textbf{0.606} & \textbf{0.553} & \textbf{0.574} & \textbf{0.568} & \textbf{0.553} & \textsc{ags}  & \textbf{0.605} & \textbf{0.554} & \textbf{0.570} & \textbf{0.567} & \textbf{0.555} \\
&       & \textsc{aks}  & \textbf{0.594} & 0.549          & \textbf{0.571} & 0.536          & \textbf{0.555} & \textsc{akgs} & \textbf{0.614} & \textbf{0.571} & \textbf{0.591} & \textbf{0.557} & \textbf{0.574} \\

\hline
\end{tabular}
}
\label{tab:medium-results}
\end{table*}

\begin{table*}[t]
\caption{F1-scores for the eight inference methods on large models. Graph-free methods (AO, AK, AOS, AKS) and graph-enhanced methods (AG, AKG, AGS, AKGS) are shown side by side. AO~=~Article Only; AK~=~Article + Keywords; AOS~=~AO + Self-Consistency; AKS~=~AK + Self-Consistency; AG~=~Article + Graph; AKG~=~AG + Keywords; AGS~=~AG + Self-Consistency; AKGS~=~AKG + Self-Consistency. Bold values exceed the baseline~\cite{sarkar2023zeroshot} (Base.\ column).}
\renewcommand{\arraystretch}{1.1}
\centering
\scriptsize
\setlength{\tabcolsep}{2pt}
\resizebox{\textwidth}{!}{%
\begin{tabular}{|l|c|l|ccccc|l|ccccc|}
\hline
\multicolumn{1}{|l|}{\textbf{}}
& \multicolumn{1}{c|}{\textbf{}}
& \multicolumn{1}{l|}{\textbf{}}
& \multicolumn{5}{c|}{\textbf{Graph-free}}
& \multicolumn{1}{l|}{\textbf{}}
& \multicolumn{5}{c|}{\textbf{Graph-enhanced}} \\
\cline{4-8} \cline{10-14}
\textbf{Dataset}
& \textbf{Base.}
& \textbf{Method}
& \shortstack{\textbf{LLaMA} \\ \textbf{3.3-70B}}
& \shortstack{\textbf{Qwen} \\ \textbf{2.5-72B}}
& \shortstack{\textbf{Qwen} \\ \textbf{3-32B}}
& \shortstack{\textbf{Gemma} \\ \textbf{3-27B}}
& \shortstack{\textbf{GPT-4o}}
& \textbf{Method}
& \shortstack{\textbf{LLaMA} \\ \textbf{3.3-70B}}
& \shortstack{\textbf{Qwen} \\ \textbf{2.5-72B}}
& \shortstack{\textbf{Qwen} \\ \textbf{3-32B}}
& \shortstack{\textbf{Gemma} \\ \textbf{3-27B}}
& \shortstack{\textbf{GPT-4o}} \\[4pt]
\hline
\multirow{4}{*}{Medical}
& \multirow{4}{*}{0.606} & \textsc{ao}   & \textbf{0.649} & \textbf{0.636} & \textbf{0.622} & \textbf{0.707} & \textbf{0.714} & \textsc{ag}   & \textbf{0.704} & \textbf{0.628} & \textbf{0.618} & \textbf{0.701} & \textbf{0.708} \\
&       & \textsc{ak}   & \textbf{0.658} & \textbf{0.699} & \textbf{0.677} & \textbf{0.759} & \textbf{0.753} & \textsc{akg}  & \textbf{0.652} & \textbf{0.702} & \textbf{0.679} & \textbf{0.759} & \textbf{0.766} \\
&       & \textsc{aos}  & \textbf{0.637} & \textbf{0.637} & \textbf{0.628} & \textbf{0.704} & \textbf{0.719} & \textsc{ags}  & \textbf{0.684} & \textbf{0.625} & \textbf{0.620} & \textbf{0.697} & \textbf{0.712} \\
&       & \textsc{aks}  & \textbf{0.644} & \textbf{0.680} & \textbf{0.661} & \textbf{0.739} & \textbf{0.736} & \textsc{akgs} & \textbf{0.619} & \textbf{0.685} & \textbf{0.663} & \textbf{0.739} & \textbf{0.738} \\
\hline
\multirow{4}{*}{News}
& \multirow{4}{*}{0.521} & \textsc{ao}   & \textbf{0.643} & \textbf{0.643} & \textbf{0.660} & \textbf{0.636} & \textbf{0.704} & \textsc{ag}   & \textbf{0.635} & \textbf{0.637} & \textbf{0.654} & \textbf{0.631} & \textbf{0.698} \\
&       & \textsc{ak}   & \textbf{0.702} & \textbf{0.713} & \textbf{0.716} & \textbf{0.687} & \textbf{0.749} & \textsc{akg}  & \textbf{0.701} & \textbf{0.711} & \textbf{0.715} & \textbf{0.689} & \textbf{0.752} \\
&       & \textsc{aos}  & \textbf{0.638} & \textbf{0.641} & \textbf{0.665} & \textbf{0.635} & \textbf{0.703} & \textsc{ags}  & \textbf{0.630} & \textbf{0.634} & \textbf{0.656} & \textbf{0.627} & \textbf{0.693} \\
&       & \textsc{aks}  & \textbf{0.674} & \textbf{0.689} & \textbf{0.696} & \textbf{0.670} & \textbf{0.719} & \textsc{akgs} & \textbf{0.677} & \textbf{0.694} & \textbf{0.699} & \textbf{0.669} & \textbf{0.721} \\
\hline
\multirow{4}{*}{Cell. phone}
& \multirow{4}{*}{0.576} & \textsc{ao}   & \textbf{0.669} & \textbf{0.647} & \textbf{0.693} & \textbf{0.664} & \textbf{0.648} & \textsc{ag}   & \textbf{0.662} & \textbf{0.637} & \textbf{0.684} & \textbf{0.658} & \textbf{0.641} \\
&       & \textsc{ak}   & \textbf{0.763} & \textbf{0.698} & \textbf{0.715} & \textbf{0.713} & \textbf{0.758} & \textsc{akg}  & \textbf{0.763} & \textbf{0.701} & \textbf{0.716} & \textbf{0.716} & \textbf{0.756} \\
&       & \textsc{aos}  & \textbf{0.688} & \textbf{0.658} & \textbf{0.681} & \textbf{0.663} & \textbf{0.676} & \textsc{ags}  & \textbf{0.677} & \textbf{0.651} & \textbf{0.672} & \textbf{0.654} & \textbf{0.668} \\
&       & \textsc{aks}  & \textbf{0.702} & \textbf{0.689} & \textbf{0.677} & \textbf{0.693} & \textbf{0.736} & \textsc{akgs} & \textbf{0.704} & \textbf{0.694} & \textbf{0.681} & \textbf{0.696} & \textbf{0.738} \\
\hline
\multirow{4}{*}{Digital cam. 1}
& \multirow{4}{*}{0.641} & \textsc{ao}   & \textbf{0.655} & \textbf{0.662} & \textbf{0.648} & \textbf{0.658} & \textbf{0.652} & \textsc{ag}   & \textbf{0.660} & \textbf{0.677} & \textbf{0.659} & \textbf{0.663} & \textbf{0.668} \\
&       & \textsc{ak}   & \textbf{0.742} & \textbf{0.714} & \textbf{0.705} & \textbf{0.687} & \textbf{0.750} & \textsc{akg}  & \textbf{0.745} & \textbf{0.715} & \textbf{0.706} & \textbf{0.685} & \textbf{0.752} \\
&       & \textsc{aos}  & 
\textbf{0.643} & \textbf{0.643} & \textbf{0.652} & \textbf{0.645} & \textbf{0.646} & \textsc{ags}  & \textbf{0.647} & \textbf{0.652} & \textbf{0.645} & \textbf{0.644} & \textbf{0.650} \\
&       & \textsc{aks}  & \textbf{0.672} & \textbf{0.701} & \textbf{0.655} & \textbf{0.663} & \textbf{0.716} & \textsc{akgs} & \textbf{0.675} & \textbf{0.701} & \textbf{0.654} & \textbf{0.665} & \textbf{0.717} \\
\hline
\multirow{4}{*}{Digital cam. 2}
& \multirow{4}{*}{0.562} & \textsc{ao}   & \textbf{0.625} & \textbf{0.638} & \textbf{0.642} & \textbf{0.663} & \textbf{0.623} & \textsc{ag}   & \textbf{0.624} & \textbf{0.623} & \textbf{0.637} & \textbf{0.653} & \textbf{0.618} \\
&       & \textsc{ak}   & \textbf{0.737} & \textbf{0.571} & \textbf{0.697} & \textbf{0.621} & \textbf{0.707} & \textsc{akg}  & \textbf{0.725} & \textbf{0.660} & \textbf{0.698} & \textbf{0.621} & \textbf{0.707} \\
&       & \textsc{aos}  & \textbf{0.631} & \textbf{0.653} & \textbf{0.638} & \textbf{0.649} & \textbf{0.644} & \textsc{ags}  & \textbf{0.627} & \textbf{0.610} & \textbf{0.632} & \textbf{0.640} & \textbf{0.632} \\
&       & \textsc{aks}  & \textbf{0.756} & \textbf{0.585} & \textbf{0.664} & \textbf{0.690} & \textbf{0.676} & \textsc{akgs} & \textbf{0.650} & \textbf{0.641} & \textbf{0.664} & \textbf{0.691} & \textbf{0.678} \\
\hline
\multirow{4}{*}{DVD player}
& \multirow{4}{*}{0.533} & \textsc{ao}   & \textbf{0.549} & \textbf{0.602} & \textbf{0.592} & \textbf{0.549} & \textbf{0.578} & \textsc{ag}   & \textbf{0.548} & \textbf{0.594} & \textbf{0.586} & \textbf{0.541} & \textbf{0.569} \\
&       & \textsc{ak}   & \textbf{0.653} & \textbf{0.692} & \textbf{0.667} & \textbf{0.600} & \textbf{0.699} & \textsc{akg}  & \textbf{0.652} & \textbf{0.692} & \textbf{0.669} & \textbf{0.599} & \textbf{0.699} \\
&       & \textsc{aos}  & \textbf{0.535} & \textbf{0.599} & \textbf{0.605} & \textbf{0.548} & \textbf{0.574} & \textsc{ags}  & \textbf{0.536} & \textbf{0.588} & \textbf{0.594} & \textbf{0.537} & \textbf{0.563} \\
&       & \textsc{aks}  & \textbf{0.586} & \textbf{0.684} & \textbf{0.699} & \textbf{0.574} & \textbf{0.698} & \textsc{akgs} & \textbf{0.586} & \textbf{0.685} & \textbf{0.699} & \textbf{0.579} & \textbf{0.699} \\
\hline
\multirow{4}{*}{Mp3 player}
& \multirow{4}{*}{0.571} & \textsc{ao}   & \textbf{0.616} & \textbf{0.577} & \textbf{0.599} & \textbf{0.613} & \textbf{0.672} & \textsc{ag}   & \textbf{0.610} & \textbf{0.608} & \textbf{0.582} & \textbf{0.607} & \textbf{0.663} \\
&       & \textsc{ak}   & \textbf{0.721} & \textbf{0.644} & \textbf{0.610} & \textbf{0.667} & \textbf{0.717} & \textsc{akg}  & \textbf{0.730} & \textbf{0.642} & \textbf{0.613} & \textbf{0.665} & \textbf{0.719} \\
&       & \textsc{aos}  & \textbf{0.615} & \textbf{0.575} & \textbf{0.580} & \textbf{0.615} & \textbf{0.668} & \textsc{ags}  & \textbf{0.607} & \textbf{0.585} & \textbf{0.574} & \textbf{0.603} & \textbf{0.659} \\
&       & \textsc{aks}  & \textbf{0.648} & \textbf{0.621} & \textbf{0.594} & \textbf{0.642} & \textbf{0.698} & \textsc{akgs} & \textbf{0.652} & \textbf{0.625} & \textbf{0.597} & \textbf{0.645} & \textbf{0.702} \\
\hline
\multirow{4}{*}{SemEval}
& \multirow{4}{*}{0.550} & \textsc{ao}   & \textbf{0.642} & \textbf{0.609} & \textbf{0.624} & \textbf{0.639} & \textbf{0.655} & \textsc{ag}   & \textbf{0.635} & \textbf{0.601} & \textbf{0.616} & \textbf{0.632} & \textbf{0.648} \\
&       & \textsc{ak}   & \textbf{0.644} & \textbf{0.626} & \textbf{0.635} & \textbf{0.638} & \textbf{0.653} & \textsc{akg}  & \textbf{0.643} & \textbf{0.625} & \textbf{0.637} & \textbf{0.640} & \textbf{0.656} \\
&       & \textsc{aos}  & \textbf{0.630} & \textbf{0.604} & \textbf{0.629} & \textbf{0.639} & \textbf{0.650} & \textsc{ags}  & \textbf{0.620} & \textbf{0.598} & \textbf{0.618} & \textbf{0.628} & \textbf{0.643} \\
&       & \textsc{aks}  & \textbf{0.638} & \textbf{0.606} & \textbf{0.621} & \textbf{0.635} & \textbf{0.647} & \textsc{akgs} & \textbf{0.637} & \textbf{0.608} & \textbf{0.621} & \textbf{0.635} & \textbf{0.649} \\
\hline
\end{tabular}
}
\label{tab:large-results}
\end{table*}

\begin{table*}[t]
\caption{Total runtime (seconds) per article across methods and model sizes.
\textsc{ag} = Article + Graph;
\textsc{akg} = AG + Keywords;
\textsc{ags} = AG + Self-Consistency (5 passes);
\textsc{akgs} = AKG + Self-Consistency. Medical and News articles are larger than those of other datasets.}
\renewcommand{\arraystretch}{1.1}
\centering
\scriptsize
\setlength{\tabcolsep}{2pt}
\resizebox{\textwidth}{!}{%
\begin{tabular}{|l|l|ccccc|ccccc|ccccc|}
\hline
\multicolumn{1}{|l|}{\textbf{}}
& \multicolumn{1}{l|}{\textbf{}}
& \multicolumn{5}{c|}{\textbf{Large models}}
& \multicolumn{5}{c|}{\textbf{Medium size models}}
& \multicolumn{5}{c|}{\textbf{Smaller models}} \\
\cline{3-17}
\textbf{Dataset}
& \textbf{Method}
& \shortstack{\textbf{LLaMA} \\ \textbf{3.3-70B}}
& \shortstack{\textbf{Qwen} \\ \textbf{2.5-72B}}
& \shortstack{\textbf{Qwen} \\ \textbf{3-32B}}
& \shortstack{\textbf{Gemma} \\ \textbf{3-27B}}
& \shortstack{\textbf{GPT-4o} \\ \textbf{}}
& \shortstack{\textbf{GPT-OSS} \\ \textbf{20B}}
& \shortstack{\textbf{Mixtral} \\ \textbf{8x7B}}
& \shortstack{\textbf{Gemma} \\ \textbf{2-9B}}
& \shortstack{\textbf{LLaMA} \\ \textbf{3.1-8B}}
& \shortstack{\textbf{Qwen} \\ \textbf{2.5-7B}}
& \shortstack{\textbf{Gemma} \\ \textbf{3n-E4B}}
& \shortstack{\textbf{LLaMA} \\ \textbf{3.2-3B}}
& \shortstack{\textbf{Qwen} \\ \textbf{2.5-3B}}
& \shortstack{\textbf{Ministral} \\ \textbf{3B}}
& \shortstack{\textbf{DS-R1} \\ \textbf{1.5B}} \\[4pt]
\hline

\multirow{4}{*}{Medical}
& \textsc{ag}  & 14.19 & 13.42 &  6.33 &  5.54 & 12.87 &  3.17 &  2.23 &  1.99 &  1.71 &  2.65 & 0.81 & 0.77 & 0.85 & 0.69 & 0.51 \\
& \textsc{akg}  & 15.63 & 14.76 &  6.92 &  6.09 & 14.18 &  3.45 &  2.41 &  2.19 &  1.93 &  2.91 & 0.89 & 0.85 & 0.93 & 0.72 & 0.56 \\
& \textsc{ags} & \textbf{70.97} & \textbf{67.03} & \textbf{31.55} & \textbf{27.61} & \textbf{64.35} & \textbf{15.79} & \textbf{11.06} & \textbf{9.87} & \textbf{8.65} & \textbf{13.26} & \textbf{3.96} & \textbf{3.92} & \textbf{4.16} & \textbf{3.37} & \textbf{2.53} \\
& \textsc{akgs} & \textbf{78.03} & \textbf{73.73} & \textbf{34.71} & \textbf{30.37} & \textbf{70.79} & \textbf{17.32} & \textbf{12.16} & \textbf{10.82} & \textbf{9.52} & \textbf{14.58} & \textbf{4.36} & \textbf{4.32} & \textbf{4.57} & \textbf{3.71} & \textbf{2.79} \\
\hline

\multirow{4}{*}{News}
& \textsc{ag}  & 14.21 & 13.38 &  6.29 &  5.50 & 12.91 &  3.13 &  2.19 &  1.95 &  1.75 &  2.61 & 0.77 & 0.81 & 0.81 & 0.65 & 0.49 \\
& \textsc{akg}  & 15.59 & 14.72 &  6.96 &  6.05 & 14.22 &  3.49 &  2.45 &  2.15 &  1.89 &  2.95 & 0.85 & 0.89 & 0.89 & 0.76 & 0.54 \\
& \textsc{ags} & \textbf{70.93} & \textbf{67.05} & \textbf{31.51} & \textbf{27.57} & \textbf{64.41} & \textbf{15.75} & \textbf{11.02} & \textbf{9.83} & \textbf{8.69} & \textbf{13.18} & \textbf{3.92} & \textbf{3.96} & \textbf{4.12} & \textbf{3.33} & \textbf{2.49} \\
& \textsc{akgs} & \textbf{78.07} & \textbf{73.69} & \textbf{34.67} & \textbf{30.33} & \textbf{70.83} & \textbf{17.36} & \textbf{12.12} & \textbf{10.86} & \textbf{9.56} & \textbf{14.54} & \textbf{4.32} & \textbf{4.36} & \textbf{4.53} & \textbf{3.67} & \textbf{2.75} \\
\hline

\multirow{4}{*}{Cellular phone}
& \textsc{ag}  & 1.44 & 1.32 & 0.65 & 0.57 & 1.31 & 0.34 & 0.24 & 0.18 & 0.19 & 0.27 & 0.10 & 0.06 & 0.10 & 0.05 & 0.04 \\
& \textsc{akg}  & 1.58 & 1.45 & 0.67 & 0.63 & 1.44 & 0.33 & 0.22 & 0.24 & 0.17 & 0.29 & 0.11 & 0.07 & 0.11 & 0.09 & 0.05 \\
& \textsc{ags} & \textbf{7.08} & \textbf{6.72} & \textbf{3.13} & \textbf{2.78} & \textbf{6.54} & \textbf{1.56} & \textbf{1.12} & \textbf{0.97} & \textbf{0.89} & \textbf{1.32} & \textbf{0.41} & \textbf{0.37} & \textbf{0.43} & \textbf{0.32} & \textbf{0.21} \\
& \textsc{akgs} & \textbf{7.82} & \textbf{7.35} & \textbf{3.49} & \textbf{3.06} & \textbf{7.17} & \textbf{1.75} & \textbf{1.19} & \textbf{1.10} & \textbf{0.93} & \textbf{1.45} & \textbf{0.45} & \textbf{0.41} & \textbf{0.48} & \textbf{0.39} & \textbf{0.23} \\
\hline

\multirow{4}{*}{Digital cam. 1}
& \textsc{ag}  & 1.40 & 1.36 & 0.61 & 0.53 & 1.29 & 0.30 & 0.20 & 0.22 & 0.15 & 0.25 & 0.06 & 0.10 & 0.06 & 0.09 & 0.04 \\
& \textsc{akg}  & 1.54 & 1.49 & 0.71 & 0.59 & 1.42 & 0.37 & 0.26 & 0.20 & 0.21 & 0.31 & 0.07 & 0.11 & 0.07 & 0.05 & 0.05 \\
& \textsc{ags} & \textbf{7.12} & \textbf{6.68} & \textbf{3.17} & \textbf{2.74} & \textbf{6.58} & \textbf{1.60} & \textbf{1.08} & \textbf{1.01} & \textbf{0.85} & \textbf{1.28} & \textbf{0.37} & \textbf{0.41} & \textbf{0.39} & \textbf{0.36} & \textbf{0.22} \\
& \textsc{akgs} & \textbf{7.78} & \textbf{7.39} & \textbf{3.45} & \textbf{3.02} & \textbf{7.21} & \textbf{1.71} & \textbf{1.23} & \textbf{1.06} & \textbf{0.97} & \textbf{1.41} & \textbf{0.41} & \textbf{0.45} & \textbf{0.44} & \textbf{0.35} & \textbf{0.24} \\
\hline

\multirow{4}{*}{Digital cam. 2}
& \textsc{ag}  & 1.43 & 1.35 & 0.62 & 0.56 & 1.33 & 0.32 & 0.23 & 0.20 & 0.17 & 0.26 & 0.09 & 0.08 & 0.08 & 0.07 & 0.04 \\
& \textsc{akg}  & 1.57 & 1.46 & 0.70 & 0.62 & 1.46 & 0.36 & 0.25 & 0.21 & 0.20 & 0.28 & 0.10 & 0.09 & 0.11 & 0.07 & 0.05 \\
& \textsc{ags} & \textbf{7.11} & \textbf{6.69} & \textbf{3.16} & \textbf{2.77} & \textbf{6.56} & \textbf{1.59} & \textbf{1.09} & \textbf{1.00} & \textbf{0.87} & \textbf{1.31} & \textbf{0.40} & \textbf{0.38} & \textbf{0.42} & \textbf{0.33} & \textbf{0.22} \\
& \textsc{akgs} & \textbf{7.81} & \textbf{7.36} & \textbf{3.48} & \textbf{3.05} & \textbf{7.19} & \textbf{1.74} & \textbf{1.20} & \textbf{1.09} & \textbf{0.94} & \textbf{1.43} & \textbf{0.44} & \textbf{0.42} & \textbf{0.47} & \textbf{0.38} & \textbf{0.24} \\
\hline

\multirow{4}{*}{DVD player}
& \textsc{ag}  & 1.41 & 1.33 & 0.64 & 0.54 & 1.30 & 0.33 & 0.21 & 0.19 & 0.16 & 0.24 & 0.08 & 0.09 & 0.09 & 0.06 & 0.03 \\
& \textsc{akg}  & 1.55 & 1.48 & 0.68 & 0.60 & 1.43 & 0.35 & 0.23 & 0.23 & 0.18 & 0.30 & 0.09 & 0.10 & 0.08 & 0.08 & 0.05 \\
& \textsc{ags} & \textbf{7.09} & \textbf{6.71} & \textbf{3.14} & \textbf{2.75} & \textbf{6.52} & \textbf{1.57} & \textbf{1.11} & \textbf{0.98} & \textbf{0.88} & \textbf{1.29} & \textbf{0.38} & \textbf{0.40} & \textbf{0.40} & \textbf{0.35} & \textbf{0.21} \\
& \textsc{akgs} & \textbf{7.79} & \textbf{7.38} & \textbf{3.46} & \textbf{3.03} & \textbf{7.15} & \textbf{1.72} & \textbf{1.22} & \textbf{1.07} & \textbf{0.96} & \textbf{1.42} & \textbf{0.42} & \textbf{0.44} & \textbf{0.45} & \textbf{0.36} & \textbf{0.23} \\
\hline

\multirow{4}{*}{Mp3 player}
& \textsc{ag}  & 1.45 & 1.34 & 0.63 & 0.55 & 1.34 & 0.31 & 0.22 & 0.21 & 0.18 & 0.28 & 0.08 & 0.08 & 0.08 & 0.07 & 0.04 \\
& \textsc{akg}  & 1.59 & 1.47 & 0.69 & 0.61 & 1.47 & 0.35 & 0.24 & 0.17 & 0.19 & 0.27 & 0.09 & 0.09 & 0.09 & 0.07 & 0.05 \\
& \textsc{ags} & \textbf{7.13} & \textbf{6.70} & \textbf{3.15} & \textbf{2.76} & \textbf{6.57} & \textbf{1.58} & \textbf{1.10} & \textbf{0.99} & \textbf{0.87} & \textbf{1.30} & \textbf{0.39} & \textbf{0.39} & \textbf{0.41} & \textbf{0.34} & \textbf{0.22} \\
& \textsc{akgs} & \textbf{7.83} & \textbf{7.37} & \textbf{3.47} & \textbf{3.04} & \textbf{7.20} & \textbf{1.73} & \textbf{1.21} & \textbf{1.08} & \textbf{0.95} & \textbf{1.44} & \textbf{0.43} & \textbf{0.43} & \textbf{0.46} & \textbf{0.37} & \textbf{0.24} \\
\hline

\multirow{4}{*}{SemEval}
& \textsc{ag}  & 1.42 & 1.36 & 0.62 & 0.56 & 1.32 & 0.32 & 0.22 & 0.20 & 0.16 & 0.26 & 0.08 & 0.08 & 0.07 & 0.07 & 0.04 \\
& \textsc{akg}  & 1.56 & 1.49 & 0.70 & 0.60 & 1.45 & 0.36 & 0.26 & 0.22 & 0.20 & 0.32 & 0.10 & 0.09 & 0.10 & 0.08 & 0.05 \\
& \textsc{ags} & \textbf{7.10} & \textbf{6.72} & \textbf{3.16} & \textbf{2.77} & \textbf{6.55} & \textbf{1.60} & \textbf{1.12} & \textbf{1.01} & \textbf{0.86} & \textbf{1.33} & \textbf{0.40} & \textbf{0.37} & \textbf{0.43} & \textbf{0.32} & \textbf{0.21} \\
& \textsc{akgs} & \textbf{7.80} & \textbf{7.39} & \textbf{3.48} & \textbf{3.05} & \textbf{7.18} & \textbf{1.75} & \textbf{1.23} & \textbf{1.09} & \textbf{0.94} & \textbf{1.46} & \textbf{0.44} & \textbf{0.41} & \textbf{0.48} & \textbf{0.36} & \textbf{0.23} \\
\hline

\end{tabular}}

\label{tab:runtime}
\end{table*}

We evaluate eight inference variants across fifteen LLMs on eight multi-label topic classification datasets. To examine how model size influences performance, we group the fifteen LLMs into three categories, small, medium, and large, with results for each group presented separately in Tables~\ref{tab:small-results}, \ref{tab:medium-results}, and \ref{tab:large-results}. In every table, each graph enhanced variant is placed side by side with its corresponding graph-free counterpart (e.g., AO with AG, and AK with AKG, AOS with AGS, and AKS with AKGS). This allows direct comparison of the impact of graph augmentation. The tables also report the baseline~\cite{sarkar2023zeroshot} (Base. column), with bold values indicating methods that outperform this baseline. Throughout this section, \textit{mean} F1-score refers to the average F1-score computed across all datasets and methods (for a model) or across all datasets and models (for a method).

\subsection{Small Models (Table~\ref{tab:small-results})}

Table~\ref{tab:small-results} presents results for five small-scale models: Gemma 3n-E4B, LLaMA 3.2-3B \cite{grattafiori2024llama}, Qwen 2.5-3B, Ministral 3B, and DeepSeek-R1-Distill-Qwen-1.5B (DS-R1-1.5B), evaluated on eight multi-label datasets.

\begin{itemize}

\item \textbf{Overall performance.} Across all model-dataset-method combinations, AKG (Article + Keywords + Graph) obtains the highest \textit{mean} F1-score (0.493). Among the evaluated small models, Gemma 3n-E4B achieves the highest overall \textit{mean} F1-score (0.521, averaged across all eight inference variants and datasets), followed by Qwen 2.5-3B (0.485) and Ministral 3B (0.476).
Small models rarely outperform the baseline \cite{sarkar2023zeroshot}. However, graph-enhanced variants exceed the baseline slightly more often than graph-free variants.

\item \textbf{Graph-free vs. graph-enhanced methods.} To isolate the contribution of knowledge graph augmentation, we compare graph-enhanced methods with their graph-free counterparts. Graph augmentation shows a consistent improvement for small-scale models, improving the \textit{mean} F1-score from 0.458 (graph-free) to 0.473 (graph-enhanced), a relative improvement of 3.26\%.

\item \textbf{Notable observations.} Surprisingly, self-consistency decoding does not improve performance and instead tends to reduce it. 
Overall, comparing the four methods without self-consistency to their self-consistency counterparts, self-consistency decoding decreases the \textit{mean} F1-score by 4.25\%.

\end{itemize}

\subsection{Medium Models (Table~\ref{tab:medium-results})}
Table~\ref{tab:medium-results} presents results for five medium-scale models: GPT-OSS-20B~\cite{agarwal2025gpt}, Mixtral 8x7B~\cite{jiang2024mixtral}, Gemma 2-9B~\cite{team2024gemma}, LLaMA 3.1-8B~\cite{grattafiori2024llama}, and Qwen 2.5-7B~\cite{yang2025qwen3}, evaluated across all eight datasets.

\begin{itemize}
\item \textbf{Overall performance.} AKG (Article + Keywords + Graph) again achieves the strongest performance among medium-scale models, obtaining the highest \textit{mean} F1-score (0.640) averaged across all model-dataset combinations. Among individual models, GPT-OSS-20B achieves the highest \textit{mean} F1-score (0.655), followed by Qwen 2.5-7B (0.623) and LLaMA 3.1-8B (0.600). 
Across all 320 evaluations (8 methods × 5 models × 8 datasets), the inference methods outperform the baseline by 7.43\% relative improvement on average. While the graph-free methods improve by 6.65\%, graph-enhanced methods improve by 8.21\%.

\item \textbf{Graph-free vs. graph-enhanced methods.} The contribution of knowledge graph augmentation is measured by comparing each graph-enhanced method with its graph-free counterpart. Graph augmentation provides a small but consistent benefit for medium-scale models, yielding an average relative improvement of 1.47\%.


\item \textbf{Notable observations.} The largest improvement over the baseline~\cite{sarkar2023zeroshot} is observed on the \emph{News} dataset with GPT-OSS-20B, where AKG (Article + Keywords + Graph) improves F1-score by 33.97\%. Another significant improvement attributable to graph augmentation occurs on the \emph{Digital Camera 2} dataset with Qwen 2.5-7B, where AKGS (Article + Keywords + Graph, self-consistency) improves over AKS (Article + Keywords, self-consistency) by 35.7\%. 



\end{itemize}

\subsection{Large Models (Table~\ref{tab:large-results})}
Table~\ref{tab:large-results} presents results for five large models: LLaMA 3.3-70B~\cite{grattafiori2024llama}, Qwen 2.5-72B, Qwen 3-32B~\cite{yang2025qwen3}, Gemma 3-27B~\cite{team2024gemma}, and GPT-4o~\cite{hurst2024gpt}, evaluated across all eight datasets.


\begin{itemize}
    \item \textbf{Overall performance.} As observed with the other model groups, AKG obtains the highest \textit{mean} F1-score (0.692), closely followed by AK (0.689). Among individual models, GPT-4o ranks first (\textit{mean} F1-score 0.685), followed by LLaMA 3.3-70B (0.655) and Gemma 3-27B (0.652). 
    All configurations outperform the baseline~\cite{sarkar2023zeroshot}, with inference methods achieving an average relative improvement of 15.37\%.
    
    \item \textbf{Graph-free vs. graph-enhanced methods.} 
    Adding the graph changes performance only marginally for large models. AKG (Article + Keywords + Graph) shows a small improvement over AK (0.42\%), while AG (Article + Graph), AGS (Article + Graph, self-consistency), and AKGS (Article + Keywords + Graph, self-consistency) show slight declines relative to their graph-free counterparts (0.39\%, 0.98\%, and 0.01\%, respectively). Overall, graph augmentation provides negligible gains for large models. This suggests that these models already capture substantial relational knowledge during pretraining.

    \item \textbf{Notable observations.} Although the average graph effect is small, the \emph{Digital Camera 2} dataset shows strong sensitivity to graph augmentation. For Qwen 2.5-72B, AKG (Article + Keywords + Graph) improves over AK (Article + Keywords) from 0.571 to 0.660, a 15.6\% increase, representing the largest positive graph effect observed in the large-model tier. In contrast, LLaMA 3.3-70B shows the largest negative graph effect, with AKGS (Article + Keywords + Graph, self-consistency) decreasing performance relative to AKS (Article + Keywords, self-consistency) by 14.0\% (from 0.756 to 0.650). Beyond \emph{Digital Camera 2}, graph augmentation also yields notable gains on other datasets: on \emph{Medical}, GPT-4o with AKG achieves an F1-score of 0.766, a 26.40\% improvement over the baseline~\cite{sarkar2023zeroshot}, and on \emph{Mp3 Player}, LLaMA 3.3-70B with AKG reaches 0.730, a 27.85\% improvement over the baseline.
    
\end{itemize}

\subsection{Runtime and Cost Analysis}
\label{sec:runtime}

As shown in Table~\ref{tab:runtime}, runtime depends on both inference technique and model size. Self-consistency variants (AOS, AKS, AGS, and AKGS) incur roughly five times more computation than single-pass approaches due to repeated inference. Among all methods, AKG offers the strongest balance of performance and computational cost. Runtime also increases with document length and model size, as evidenced by datasets such as \emph{Medical} and models such as GPT-4o. Evaluating F1-score per second further highlights the efficiency of smaller models: GPT-OSS-20B achieves competitive performance with substantially lower inference time. Qwen 2.5-7B and Gemma 3-27B also provide favorable accuracy-cost trade-offs. 

\subsection{Interesting Findings}
\label{sec:summary-findings}
\begin{itemize}
    \item \textbf{AKG is the strongest inference strategy overall.} The AKG (Article + Keywords + Graph) configuration achieves the highest \textit{mean} F1-score and is the best-performing method across all fifteen models, demonstrating the combined value of document context, keyword guidance, and graph knowledge.

    \item \textbf{Graph augmentation improves precision at the cost of recall for smaller and medium-scale models.} For small and medium models, graph-enhanced methods increase average precision by 13.2\% while reducing recall by 9.65\%. This yields a net gain in F1-score. This pattern is illustrated by Qwen 2.5-7B on a \textit{Digital Camera} sample with ground-truth labels \textit{camera}, \textit{picture}, and \textit{size}: AK (Article + Keywords) predicts an additional irrelevant label (\textit{focus}) while missing the label \textit{camera}. With graph augmentation, AKG (Article + Keywords + Graph) removes the incorrect \textit{focus} prediction and correctly identifies \textit{camera}, but misses the valid \textit{size} label. This illustrates how graph guidance improves precision through more selective, but occasionally incomplete, predictions.

    \item \textbf{Graph augmentation can introduce noise for large models.} While graphs improve performance in many cases, they can occasionally degrade it for large models by introducing irrelevant associations that encourage over-prediction, increasing false positives and reducing precision. For example, on a \textit{Cellular phone} sample with ground-truth label \textit{phone}, LLaMA 3.3-70B under AO (Article Only) correctly identifies the relevant topic, whereas AG (Article + Graph) introduces an unrelated additional label (\textit{feature}). This demonstrates how graph associations can sometimes increase false positives.

    \item \textbf{Self-consistency decoding raises computational cost without improving performance.} Surprisingly, self-consistency decoding reduces the \textit{mean} F1-score by 2.6\% across all model sizes. While 3-of-5 majority voting improves \textit{mean} precision by 9.7\%, it reduces \textit{mean} recall by 14.23\%. This indicates that it filters out valid predictions along with uncertain ones. 

    \item \textbf{Model scale matters, but parameter count alone does not determine performance.} Large models generally achieve higher performance, but architecture and training also play a substantial role. GPT-4o attains the highest overall \textit{mean} F1-score (0.685), followed by GPT-OSS-20B (0.655) and LLaMA 3.3-70B (0.655). Notably, GPT-OSS-20B, despite its medium-scale classification, outperforms three large models, including Qwen 2.5-72B, Gemma 3-27B, and Qwen 3-32B. This shows that a larger parameter count does not always guarantee stronger zero-shot classification performance.
\end{itemize}

%% file: latex/Sections/conclusion.tex
\section{Conclusion}
\label{sec:conclusion}

This work investigated whether document level knowledge graphs can improve zero-shot multi-label topic classification, addressing the common practice of representing documents and labels as flat, unstructured text. We proposed a knowledge graph-enhanced inference framework that constructs a per-article knowledge graph using an LLM driven pipeline and supplies it to the LLM alongside the article text at inference time, without requiring external knowledge bases, labeled data, or task specific fine tuning. Through an exploratory study spanning fifteen LLMs, eight datasets, and eight inference variants, we find that the benefit of explicit relational knowledge depends strongly on model scale: graph augmentation consistently improves smaller and medium scale models, primarily through gains in precision, and in several cases allows medium sized models to match or outperform larger ones, while offering negligible effects for large models, which already encode much of this knowledge from pretraining. We further find that self-consistency decoding offers no consistent performance gains despite a roughly fivefold increase in inference cost.

These findings offer a more nuanced, scale dependent picture of when structured relational knowledge is useful, a perspective not yet established in the zero-shot classification literature. They also motivate future work on mitigating graph induced noise in large models, exploring alternative structured representations, and studying how graph quality mediates downstream performance.

%% file: latex/Sections/acknowledgement.tex
\section*{Acknowledgment}

This work was supported by the NSF AI Institute for Foundations of Machine Learning.

%% file: latex/Sections/bib.tex
\vspace{-1.4cm}
\begin{IEEEbiography}[{\includegraphics[width=1in,height=1.25in,clip,keepaspectratio]{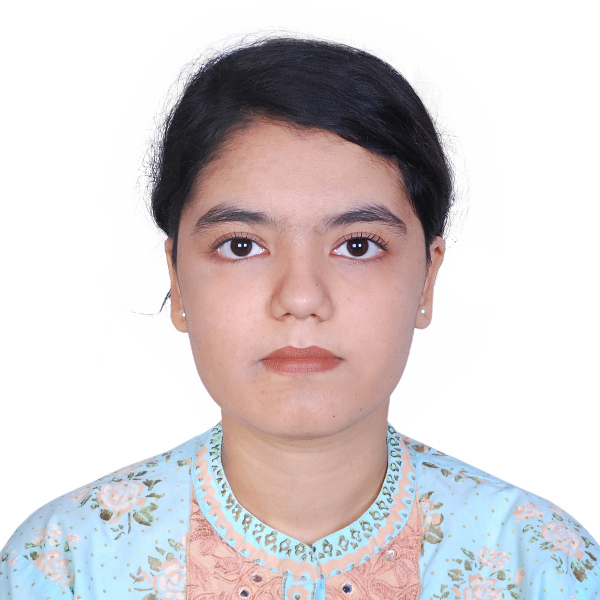}}]{Shahana Akter} is a Ph.D. student in the School of Computing at Wichita State University, where she works in the Accessible AI lab. She earned her B.Sc. in Computer Science and Engineering from Khulna University of Engineering \& Technology (KUET), in 2025. Her research investigates large language models (LLMs) with emphasis on real-world evaluation and deployment. Her primary research focus is on improving the robustness and usability of artificial intelligence systems in real-world settings.
\end{IEEEbiography}

\vspace{-1.5cm}
\begin{IEEEbiography}
[{\includegraphics[width=1in,height=1.25in,clip,keepaspectratio]{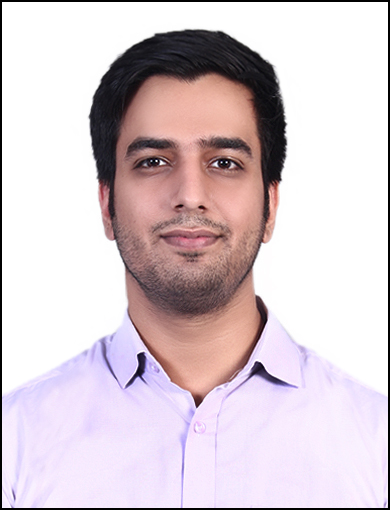}}]{Yatharth Vohra} is an M.S. student in the Department of Computer Science and Engineering at the University of Nevada, Reno, NV, USA. He received the bachelor’s degree in Computer Science and Engineering from Manav Rachna University, Faridabad, India, in 2022. His research interests include large language models (LLMs), retrieval-augmented generation (RAG), natural language processing, and generative artificial intelligence. His work focuses on developing reliable and effective LLM-based systems for real-world applications.
\end{IEEEbiography}

\vspace{-1.2cm}

\begin{IEEEbiography}
[{\includegraphics[width=1in, clip, keepaspectratio]{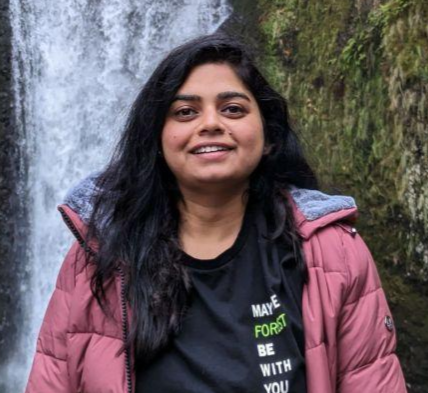}}]{Ankita Shukla} (Member, IEEE) received the master’s and Ph.D. degrees in ECE from the Indraprastha Institute of Information Technology Delhi (IIIT-Delhi), New Delhi, India. She was a Postdoctoral Researcher with Arizona State University, Tempe, AZ, USA. She is currently an Assistant Professor with the Department of Computer Science and Engineering, University of Nevada, Reno, NV, USA. Her research interests include machine learning, natural language processing, large language models, knowledge-enhanced learning, multimodal learning, and geometric and topological representation learning. Her work focuses on developing reliable and efficient artificial intelligence methods for language, vision, time-series, and scientific data, with applications in scientific discovery, healthcare, and social good.
\end{IEEEbiography}
\vspace{-1.2cm}
\begin{IEEEbiography}[{\includegraphics[width=1in, clip, keepaspectratio]{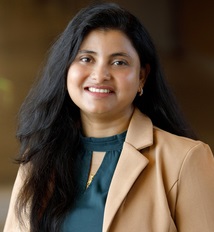}}]{Souvika Sarkar (M'25)} received her Ph.D. in Computer Science and Software Engineering from Auburn University in 2020. She is currently an Assistant Professor in the School of Computing at Wichita State University, Wichita, KS, USA. Prior to that, she completed her B.Tech in Computer Science and Engineering from West Bengal University of Technology, India, in 2012, and earned her M.E. in Software Engineering from Jadavpur University in 2018. Dr. Sarkar’s research interests include natural language processing (NLP), generative AI, LLMs, and building scalable and adaptive AI systems. Her research centers on developing efficient NLP systems, with a strong focus on interpretability, adaptability, and real-world impact. She has published widely in top-tier venues in AI and NLP. She actively serves the research community as a reviewer for leading conferences, including ACL, EMNLP, NAACL, and AACL.
\end{IEEEbiography}